\documentclass[review]{elsarticle}
\usepackage{lineno,hyperref}
\usepackage{epsfig}
\usepackage{graphicx}
\usepackage{amsmath}
\usepackage{amssymb}
\usepackage[ruled]{algorithm2e} 
\usepackage{subfig}
\usepackage{amsthm}
\usepackage{algorithmic}
\usepackage{amssymb}
\usepackage{multirow} 
\usepackage[utf8]{inputenc}
\DeclareUnicodeCharacter{FB01}{fi}
\DeclareUnicodeCharacter{FF0C}{fi}

\modulolinenumbers[5]

\begin{document}

\begin{frontmatter}

\title{Multi-view Subspace Clustering via Partition Fusion}

\author[UESTC]{Juncheng~Lv}
\author[UESTC]{Zhao~Kang}
\ead{zkang@uestc.edu.cn}
\author[UWO]{Boyu~Wang}
\ead{bwang@csd.uwo.ca}
\author[UESTC]{Luping Ji}
\ead{jiluping@uestc.edu.cn}
\author[UESTC]{Zenglin Xu}
\ead{zlxu@uestc.edu.cn}

\address[UESTC]{School of Computer Science and Engineering, University of Electronic Science and Technology of China, Chengdu, China}
\address[UWO]{Department of Computer Science, University of WesternOntario, Canada}


\begin{abstract}
Multi-view clustering is an important approach to analyze multi-view data in an unsupervised way. Among various methods, the multi-view subspace clustering approach has gained increasing attention due to its encouraging performance. Basically, it integrates multi-view information into graphs, which are then fed into spectral clustering algorithm for final result. However, its performance may degrade due to noises existing in each individual view or inconsistency between heterogeneous features. Orthogonal to current work, we propose to fuse multi-view information in a partition space, which enhances the robustness of Multi-view clustering. Specifically, we generate multiple partitions and integrate them to find the shared partition. The proposed model unifies graph learning, generation of basic partitions, and view weight learning. These three components co-evolve towards better quality outputs. We have conducted comprehensive experiments on benchmark datasets and our empirical results verify the effectiveness and robustness of our approach.
\end{abstract}

\begin{keyword}
\texttt{Multi-view learning}\sep \texttt{subspace clustering}\sep \texttt{Partition space}\sep \texttt{Information fusion}
\end{keyword}

\end{frontmatter}


\section{Introduction}
\label{sec:introduction}
In many real-world problems, data are collected from different sources in diverse domains or described by various feature collectors \cite{ding2018robust,yin2018multiview,zhang2015low,zhang2019joint,wang2018multiview}. For instance, human activities can be captured by RGB video cameras, depth cameras, or on-body sensors \cite{li2016multi,yao2019multi};
pictures shared on websites are usually surrounded by textual tags and descriptions; the same news is reported in various countries with different languages. To process these kinds of data, a number of multi-view learning algorithms have been developed \cite{xu2013survey,sun2013survey,tang2018consensus,hu2020multi,zhan2018multiview}. These methods are based on two fundamental assumptions: consistency and complementarity across views. That is to say, there exists consistent information which is shared by all views and complementary knowledge that is not contained in all views. Therefore, it is crucial for learning algorithms to make good use of the multi-view information. For example, instead of relying on a single view, multi-view clustering, attempts to achieve a better performance by integrating compatible 
and complementary information from different views \cite{huang2015spectral,blaschko2008correlational,chaudhuri2009multi,wang2015robust}. 

In recent years, plenty of multi-view clustering techniques have been proposed \cite{chao2017survey,zhan2017graph,liu2019efficient,zhuge2019simultaneous,liu2018late,wang2015multi}. They can be roughly divided into the following categories. First, the co-training style algorithms intend to maximize the mutual agreement across all views and arrive at their broadest consensus \cite{bickel2004multi,kumar2011co,kumar2011core}. Second, kernel-based methods use pre-defined kernels corresponding to different views and then combine these kernels either linearly or non-linearly in order to improve the clustering performance\cite{de2010multi,yu2012optimized,lu2014multiple,tzortzis2012kernel}. Third, graph-based methods have been derived from traditional spectral clustering with the help of some similarity measures\cite{tang2009clustering,hussain2014multi,nie2016parameter,zhan2018multiview}. Fourth, subspace clustering based methods. Subspace clustering tries to find underlying subspaces such that all data points can be segmented correctly and each group fits into one of the low-dimensional subspaces \cite{vidal2011subspace,liu2012robust,chen2012fgkm,elhamifar2009sparse,peng2018structured,liu2019robust,li2018geometric}. Due to its promising performance, varieties of multi-view subspace clustering methods have been proposed in the last few years \cite{gao2015multi,zhang2017latent,tang2018learning}. 

Most multi-view subspace clustering methods learn the sample affinity graph matrix of each view by deploying features of different views, and then build a consensus graph $S$ \cite{gao2015multi,zhang2017latent,wang2016iterative}. Some other approaches directly learn a common graph matrix $S$ \cite{abavisani2018multimodal}. Subsequently, spectral clustering algorithm \cite{ng2002spectral,Kang2018unified} is implemented on the graph Laplacian constructed by $S$ to obtain the final clustering result. Therefore, these methods ahopt a two-steps procedure. Since data are often noisy or corrupted in practice, one critical issue with these approaches is that the learned graph is often noisy, inaccurate, and fails to reveal the true relationships between data points \cite{kang2019robust,zhang2019robust}. In return, this fixed graph will deteriorate the downsteam clustering task, which makes the entire learning procedure suboptimal.

Orthogonally to integrating the multi-view information into a single graph, in this paper, we propose to fuse partitions to improve the model robustness. Note that the underlying assumption for multi-view clustering is that there exists a unique clustering pattern shared by all views. Hence, the partition space should be more robust to noise. On the one hand, even if the graphs are contaminated, the cluster structure might be slightly influenced or even remain intact. On the other hand, even if some of the partitions are severely damaged, one could still obtain a reasonable performance based on our partition fusion technique. In specific, we adaptively learn a weight for each view to control its contribution to the final clustering. The final clustering is achieved through a purposely designed weighting mechanism imposed on basic partitions.
 

 In summary, the main contributions of this work are three-fold:
 \begin{enumerate}

     \item{We propose to fuse multi-view information in a partition space. A novel fusion mechanism is further developed to find the consensus clustering.}
     
     \item{We present a unified multi-view subspace clustering model which iteratively learns a graph for each view, a partition for each view, a weight for each pratition, and a consensus clustering. By leveraging the inherent interactions between these four subtasks, they enhance each other. }
     \item{Extensive experiments on benchmark data sets validate the effectiveness of our model. Experiments on noisy data demonstrate the robustness of our approach.}
  
 \end{enumerate}
 The rest of this paper is organized as follows. After a review of the
related work in Section \ref{relate}, we introduce our proposed framework
for multi-view subspace learning in Section \ref{propose}. In Section \ref{experiment},
we evaluate the clustering performance of our algorithm. The robustness evaluation of the proposed framework on noisy data sets is presented in Section \ref{noise}. Finally,
we conclude the paper in Section \ref{conclude}.

\textbf{Notation Summary}
In this paper, we represent the matrix with capital letter and vector with lower case letter. For a matrix $A\in \mathbb{R}^{m\times n}$, $A_{i,:}$ and $A_{:,j}$ represent the $i$-th row and $j$-th column of $A$, respectively. The $\ell_2$-norm of vector $x$ is denoted by $\|x\|_2=\sqrt{x^\top \cdot x}$, where $\top$ stands for the transpose operation. The trace operator is written as $Tr(\cdot)$. The definition of Frobenius norm of $A$ is $\|A\|_F=\sqrt{\sum_{i=1}^m\sum_{j=1}^n A_{ij}^2}$. $A\geq 0$ indicates all elements of $A$ are nonnegative. $I$ is the identity matrix with a proper size.   

\section{Related Work}
\label{relate}

Multi-view Subspace Clustering (MVSC) is developed on the basis of subspace clustering. We denote the multi-view data $X$ with $[X^1,X^2,\cdots,X^t]\in \mathbb{R}^{m\times n}$, where $X^v\in \mathbb{R}^{m_v\times n}$ represents the $v$-th view data with $m_v$ features and $m=\sum_v m_v$. Basically, \cite{gao2015multi,cao2015diversity} propose to learn a graph on individual view by solving 
\begin{equation}
\min_{S^v} \sum\limits_{v=1}^t \|X^v-X^vS^v\|_F^2+\alpha f(S^v)\hspace{0.15cm} s.t.\hspace{0.15cm}S^v\ge 0,
\label{msc}
\end{equation}
where $S^v\in \mathbb{R}^{n\times n}$ denotes the graph for the $v$-th view, $f(\cdot)$ represents a certain regularizer function, and $\alpha$ is a regularization parameter to balance the model complexity and the fitting loss. $S^v$ is also called the self-expression coefficient matrix, which expresses each sample as a linear combination of other samples. Hence, it can measure the similarities between samples. Based on it, subspace clustering implements spectral clustering algorithm \cite{ng2002spectral}.

In \cite{gao2015multi}, the authors enforce that all graphs share a unique cluster indicator matrix, i.e.,
\begin{equation}
   \min_{F} \sum\limits_v Tr(F^\top L^vF) \quad s.t.\quad F^\top F=I,
   \label{multi}
\end{equation}
where graph Laplacian $L^v=D^v-S^v$ with diagonal matrix $D^v$ defined as $d_{ii}^v=\sum_j s_{ij}^v$, $F\in\mathbb{R}^{n\times c}$ is the cluster indicator matrix, and $c$ is number of clusters. It is obvious that $F$ is negotiated across different graphs, i.e., different $L^v$'s correspond to the unique $F$. Therefore, the final $F$ might not be optimal.

 Different from the above approach, the authors in \cite{wang2016iterative,cao2015diversity} first calculate the consensus graph $S$ based on averaging, i.e., $\sum\limits_{v=1}^t S^v/t$. Then, spectral clustering is applied. This simple approach fails to distinguish the different contributions of various views. Afterwards, we can obtain the final cluster indicators by performing $k$-means on $F$.

Another class of methods just consider the consistent graph of all views and their objective function can be written as \cite{abavisani2018multimodal,zhuge2017robust}
\begin{equation}
\min_{S} \sum\limits_{v=1}^t  \|X^v-X^vS\|_F^2+\alpha \mathcal{R}(S)\hspace{0.15cm} s.t.\hspace{0.15cm}S\ge 0,
\end{equation}
where $\mathcal{R}(Z)$ is some regularization function which varies in different algorithms. For instance, researchers in \cite{abavisani2018multimodal} use popular low-rank and sparse regularizers simultaneously. It is easy to see that just one graph can not preserve the flexible local mannifold structure of different views \cite{wang2016iterative}. 

We can observe that above approaches integrate multi-view information in the data space and their performance totally depends on the quality of graph. Once the graph is fixed, the rest of MVSC process reduces to spectral clustering, which is not subject to change. In real-world applications, data are often contaminated due to noise or outliers, which deteriorates the resulting graph. Consequently, the clustering performance will degrade. Therefore, we argue that direct manipulation on graphs might not be a good idea to make use of multi-view information.

Instead, we propose to integrate multi-view knowledge by fusing partitions. Concretely, we generate one partition for each view, which forms the basic partitions. Then, we seek for a consensus clustering from them based on our purposely designed weighting mechanism. This change in operation domain accompanies a number of advantages. First, each partition will capture the intrinsic cluster structure, so it is easy to find an agreement among all partitions. Second, even if one partition is heavily distorted, its contribution can be reduced by assigning a small weight, so as to prevent it from adversely affecting consensus clustering. As a result, this approach could be robust to noise. 
 
Note that the proposed approach is different from ensemble clustering \cite{wu2014k,tao2017ensemble} in several aspects. First,  only one partition is generated for each view. On the contrary, many partitions are often produced for ensemble clustering. Second, the basic partitions are integrated differently. For example, Tao et al.\cite{tao2017ensemble} adopt a low-rank and sparse decomposition strategy to discover the connection among views and detect noises. Third, a unified framework is utilized, i.e., the generation of basic partitions and their integration are joint performed. In contrast, basic partitions are input of ensemble clustering.

\section{Proposed Methodology}
\label{propose}
  In this section, we will introduce our novel model and its solution. 
\subsection{Formulation}
As in aforementioned MVSC \cite{gao2015multi,guo2015robust}, we combine graph construction and spectral clustering together. Different from it, we generate one partition for each view. Multiple partitions allow us to manipulate multi-view information in a partition space, which enhances the model robustness. In specific, we have
\begin{equation}
    \begin{split}
        \min_{S^v,F_v} \sum\limits_{v=1}^t \!\|X^v-X^vS^v\|_F^2+\!  \alpha Tr(F_v^\top L^v F_v)+\beta\|S^v\|_F^2 \\
        s.t.\quad F_v^\top F_v=I,\hspace{.1cm} S^v\ge 0,
         \label{original_method}
    \end{split}
\end{equation}
where $F_v\in \mathbb{R}^{n\times c}$ is the partition result for view $v$. We can see that Eq. (\ref{original_method}) will provides multiples partitions. For multi-view clustering, all views share a unique cluster pattern. Due to noises in defferent views or the heterogeneity of features, these partition matrices are not identical in general. Therefore, our next goal is to find a consensus clustering from them. 

To treat each view discriminatively, we propose a weighting mechanism.
To this end, we introduce a variable $w_v$ for view $ v$ which characterizes the importance of view $v$. Larger value means more contribution of this partition to consensus clustering. It is reasonable to propose the following objective function
 \begin{equation}
     \begin{split}
         \min_{w_v,S^v,F_v} \sum\limits_{v=1}^t w_v\! \Big\{\|X^v\! -\!X^vS^v\|_F^2\!+\!\alpha Tr(F_v^\top L^v F_v)\!+\! \beta\|S^v\|_F^2\Big\}\\
s.t.\hspace{.1cm} \sum\limits_v w_v=1,\hspace{.1cm}  w_v \ge 0,\hspace{.1cm}  F_v^\top F_v=I, \hspace{.1cm}  S^v \ge 0
         \label{weight}
     \end{split}
 \end{equation}
If the loss in the bracket is small, weight $w_v$ will have a large value; vice versa. 

After we obtain the weight for each view, how can we reach the consensus clustering, i.e., a partition that fits all views?
Unlike classification or regression, the cluster indicator matrix for each view is not unique. In general, for each unique clustering with $c$ clusters, there are $c!$ ($c$ factorial) equivalent representations \cite{tao2018reliable}. So the popular Euclidean distance is not applicable to measure the distance between indicator matrices any more and we need to figure out how to define the distances between partitions. 

To address above challenge, we propose to use inner product $F_vF_v^\top$, which actually represents the similarities among all data points in the $v$-th view. It is easy to understand that it is invariant with respect to $c!$ permutations. Hence, we can measure the partition distance in terms of $F_vF_v^\top$. Specifically, we develop the following partition fusion objective function
\begin{equation}
    \begin{split}
      \min_{Y\in \mathbb{R}^{n\times c}, Y^\top Y=I}\|YY^\top-\sum\limits_{v=1}^t w_vF_vF_v^\top\|_F^2,
      \label{fuse}
    \end{split}
\end{equation}
where $Y$ is the consensus cluster indicator matrix. If $w_v$ has a large value, the $v$-th partition will contribute a lot to final $Y$.

Finally, we can combine Eqs. (\ref{weight}) and (\ref{fuse}) to form a unified objective function, so that each variable can be iteratively updated. Our proposed Partition Fusion multi-view Subspace Clustering (PFSC) can be formulated as 
\begin{equation}
   \begin{split}
       \min_{S^v,F_v,w_v,Y} & \sum\limits_{v=1}^t w_v\Big\{\|X^v-X^vS^v\|_F^2+ \alpha Tr(F_v^\top L^v F_v)+\beta\|S^v\|_F^2\Big\}\\
&\hspace{.3cm}+\|YY^\top-\sum\limits_{v=1}^t w_vF_vF_v^\top\|_F^2 \\
        s.t.& \hspace{.05cm} \sum\limits_{v=1}^t w_v=1,\hspace{.05cm} w_v \ge 0,\hspace{.05cm} S^v \ge 0
        \\
        &\hspace{.05cm} F_v^\top F_v=I,\hspace{.05cm} Y^\top Y=I.
       \label{final_method}
   \end{split}
\end{equation}\label{obj}
With input data $X$, PFSC will output cluster indicator matrix $Y$. Hence, it is an end-to-end clustering method. Compared to existing work in the literature, it enjoys the following properties.
\begin{itemize}
    \item Orthogonal to existing multi-view clustering methods, our proposed model integrates multi-view information in a partition space. Since all partitions admit a unique cluster pattern, it is natural to implement information fusion in a partition space.
    \item A weight is dynamically learned for each view, which can identify the importance of each view.
    \item This uniﬁed framework seamlessly integrates the graph learning, spectral clustering, weight learning, and partition fusion. By iteratively updating $S, F, w, Y$, they can be repeatly improved. This joint learning strategy facilitates to obtain a better solution.
\end{itemize}

\subsection{Optimization}

Since Eq.(\ref{final_method}) involves several coupled variables, it is difficult to solve it. Therefore, we divide the original problem into four subproblems and develop an alternating and iterative algorithm. $S^v$, $F_v$, $w_v$, $Y$ can be solved effectively and individually by fixing the others. 

\textbf{$S^v$-subproblem}: By fixing $F_v$, $w_v$, and $Y$, we update $S^v$ by solving:
\begin{equation}
       \min_{S^v}  \sum_v \Big\{\|X^v-X^vS^v\|_F^2+\alpha Tr(F_v^\top L^v F_v)+\beta \|S^v\|_F^2\Big\}.
\end{equation}
Note that $S^v$'s are independent for each view, hence we can solve them separately. For convenience, we ignore the subscript/superscript tentatively and get
\begin{equation}
    \min_{S}Tr(-2X^\top XS+S^\top X^\top XS)+\alpha Tr(F^\top LF)+\beta Tr(S^\top S).
    \label{solves}
\end{equation}
Setting the the derivative of Eq. (\ref{solves}) with respect to $S$ to zero, and noting that $\sum_{i,j}\frac{1}{2}\|F_{i,:}-F_{:,j}\|_2^2s_{ij}=Tr(F^\top LF)$, we obtain the analytical solution for $S_{:,i}$,
 \begin{equation}
     S_{:,i}=(X^\top X+\beta I)^{-1}(X^\top X_{:,i}-\frac{\alpha}{4}d_i),
     \label{updates}
 \end{equation}
where $d_i\in\mathbb{R}^{n\times 1}$ is a vector with the $j$-th element as $d_{ij}=\|F_{i,:}-F_{j,:}\|_2^2$.
Note that once parameter $\alpha$ is given, the inverse is fixed in each iteration. In other words, we can calculate it in advance to save computing time. After obtaining $S$, we can set its negative elements to zero to ensure its nonegative.
 
 \textbf{$F_v$-subproblem}: Omitting all other unrelated terms with respect to $F_v$, we obtain
 \begin{equation}
     \min_{F_v,F_v^\top F_v=I}\alpha \sum\limits_v w_vTr(F_v^\top L^vF_v)+\|YY^\top-\sum_{v}w_vF_vF_v^\top\|_F^2.
 \end{equation}
 Similarly, $F_v$ can  be solved separately for each view. Then, we have:
 \begin{equation}
     \min_{F_v,F_v^\top F_v=I}Tr(F_v^\top MF_v),
      \label{updatef}
 \end{equation}
where $M=\alpha L_v+w_vI-2YY^\top-2 \sum_{j\ne v} F_jF_j^\top$. Then, the optimal solution $F_v$ is the $c$ eigenvectors of $M$ corresponding to the $c$ smallest eigenvalues.
 
  \textbf{$w_v$-subproblem}: Our problem can be simplified as
  \begin{equation}
  \begin{split}
      \min_{} w^\top Pw-qw  \\
      {\rm s.t.} \quad\sum_{v}w_v=1, \quad w_v\ge 0,
  \end{split}
  \label{updateW}
  \end{equation}
where $P\in\mathbb{R}^{n\times n}$ with $P_{ij}=Tr[F_iF_i^\top\times F_jF_j^\top]$ and $q$ is a vector with
 \begin{equation}
 \begin{split}
    & q_i=-g_i+2Tr(YY^\top F_iF_i^\top),\\
     & g_i=\|X^i-X^iS^i\|_F^2+\alpha Tr(F_i^\top L^iF_i)+\beta \|S^i\|_F^2.
 \end{split}
 \end{equation}
It is a standard quadratic programming problem, which can be solved efficiently.

  \textbf{$Y$-subproblem}: As for $Y$, we have the following equivalent formulation
  \begin{equation}
      \min_{Y,Y^\top Y=I} Tr[Y^\top (I-2\sum_v w_vF_vF_v^\top)Y],
      \label{updateY}
  \end{equation}
 which can be solved by singular value decomposition (SVD).
 
 
The entire optimization procedure for (\ref{final_method}) is summarized in \textbf{Algorithm 1}\footnote{Our source code will be released later.}. The main computational burden lies in the matrix inversion (line 2) and SVD (line 6). Since we just consider the top $c$ eigenvectors, the complexity of SVD is $O(n^2c)$. Therefore, the overall time complexity of Algorithm 1 is $O(n^3)$. 

\begin{algorithm}
\caption{Optimization for PFSC}
\label{alg1}
 {\bfseries Input:} multi-view data matrix $X^1, \cdots, X^t$, cluster number $c$, parameters $\alpha$, $\beta$.\\
{\bfseries Output:} $S^v, F_v, w_v,Y$.\\
{\bfseries Initialize:} Random matrix $F_v$, $w_v=1/t$.\\
 {\bfseries REPEAT}
\begin{algorithmic}[1]
\STATE \textbf{for} view 1 to $v$ \textbf{do}
\STATE  Update each column of $S$ according to (\ref{updates});\\
\STATE  S=max(S,0);\\
\STATE Solve the subproblem (\ref{updatef});\\
\STATE  Update $w_v$ via (\ref{updateW}).\\
\STATE  \textbf{end for}
\STATE   Solve the subproblem (\ref{updateY}).
\end{algorithmic}
\textbf{ UNTIL} {stopping criterion is met.}
\end{algorithm}

  \begin{table*}[!hbtp]
\begin{center}
\renewcommand{\arraystretch}{0.8}
\caption{Description of the datasets. The feature number for each view is shown in parenthesis. \label{data}}
\label{datasets} \scalebox{0.78}{
\begin{tabular}{clll}
\hline
{View} &{HW} & {Caltech7} & {Caltech20}  \\\hline
1& Profile correlations (216) & Gabor(48) & Gabor (48)\\
2& Fourier coefficients (76) &Wavelet moments (40) & Wavelet moments (40)\\
3& Karhunen coefficients (64) &CENTRIST (254)  &CENTRIST (254)\\
4 &  Morphological (6) & HOG (1984) &HOG (1984)\\
5& Pixel averages (240) &GIST (512) & GIST (512) \\
6& Zernike moments (47) & LBP (928) &LBP (928)\\\hline
Data points & 2000 & 1474 & 2386\\
Classes& 10 & 7  & 20 \\
\hline
\end{tabular}}
\end{center}
\end{table*}

\section{Experiments on Benchmark Data}
 \label{experiment}
\subsection{Data sets}
 To fully assess the effectiveness of our proposed method, we conduct experiments on three widely used data sets with six types of features. 
 
 \textbf{Handwritten numerals (HW)} \footnote{http://archive.ics.uci.edu/ml/datasets.html.} data set is selected from UCI machine learning repository. It is comprised of 2000 data points for 0 to 9 digit classes, and each class has 200 data points. 
 
\textbf{Caltech7} \footnote{http://www.vision.caltech.edu/Image\_Datasets/Caltech101/.} is an object recognition data set with 7 categories, including “Face, Snoopy, Garfield, Motorbikes, Dolla-Bills, Stop-sign, and Windsor-chairs”. 

\textbf{Caltech20} contains 20 classes ''Brain, Camera, Face, Ferry, Rhino, Pagoda, Snoopy, Wrench, Stapler, Leopards, Hedgehog, Garfield, Binocular, Motorbikes, Windsor Chair, Car-Side, Dolla-Bill, Stop-Sign, Yin-yang, and Water-Lilly''. 

The specific characteristics of these data sets are summarized in Table \ref{data}.

 \begin{figure*}[!htbp]
\includegraphics[scale=0.26]{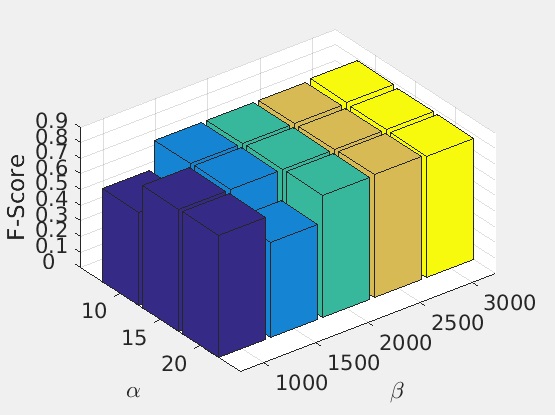}
\includegraphics[scale=0.26]{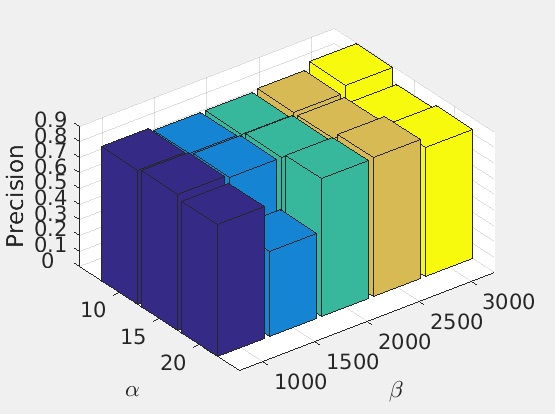}
\includegraphics[scale=0.26]{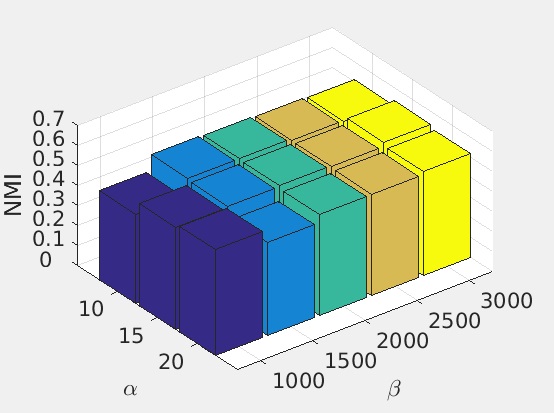}
\caption{Sensitivity analysis of parameters for our method over Caltech7 dataset evaluated with F-Score, Precision, and NMI.} \label{fig1}
\end{figure*}

\subsection{Comparison Methods}
 To evaluate the performance of the presented method, we compare it with several state-of-the-art clustering methods.
 \begin{itemize}
\item{The classic $k$-means clustering algorithm (KM): The KM clustering on concatenated features is used as a baseline algorithm. }
\item{Co-regularized multi-view spectral clustering (Co-reg)~\cite{kumar2011core}: A co-regularization mechnism is utilized to ensure that partitions from different views are close to each other.}
\item{Co-trained multi-view spectral clustering (Co-train)~\cite{kumar2011co}: A co-training approach is used to learn multiple Laplacian eigenspace.}
\item{Multi-view kernel $k$-means clustering (MKKM)~\cite{tzortzis2012kernel}: In MKKM, data are first mapped into high-dimensional space by kernel trick. Then, kernels from different views are combined based on a weighting principle.}
\item{Robust multi-view $k$-means clustering (RMKM)~\cite{cai2013multi}: To cope with outliers, this method adopts structured sparsity-inducing norm to integrate multi-view information. }
\item{Multi-view subspace clustering (MVSC) \cite{gao2015multi}: This method simultaneously learns multiple graphs and forces them generate the same cluster structure. }
\item{Multi-manifold regularized nonnegative matrix factorization (MNMF)~\cite{zong2017multi}: Based on NMF, this method preserves the local geometrical structure of multi-view data.}
\item{Auto-weighted multiple graph learning (AMGL) \cite{nie2016parameter}: Different from Eq. (\ref{multi}), a weight is assigned for each view. }
\item{Multi-View Ensemble Clustering (MVEC) \cite{tao2017ensemble}: Solve multi-view clustering in an ensemble clustering way. }
\item{Multiple Partitions Aligned Clustering (mPAC) \cite{kang2019multiple}: This recent method uses a different approach to combine basic partitions.}
\end{itemize}

  \begin{table}[t]
\begin{center}
\renewcommand{\arraystretch}{1.1}
\caption{Clustering performance on Caltech7 data. }
\label{Caltech7} \scalebox{0.75}{
\begin{tabular}{|c|c|c|c|c|c|c}
\hline
{Method} & {F-Score}& {Precision} & {Recall}&{NMI}&{Adj-RI}\\
\hline

{KM}&0.4688(0.0327)& \textbf{0.7868(0.0080)}&0.3618(0.0371)&0.4278(0.0120)&0.3172(0.0297)\\
\hline
{Co-train}&0.4678(0.0172)&0.7192(0.0136)&0.3550(0.0168)&0.3235(0.0226)&0.3342(0.0157)\\
\hline
{Co-reg}& 0.4981(0.0092)&0.7014(0.0076)&0.3622(0.0098)&0.3738(0.0061)&0.2894(0.0046) \\
\hline
{MKKM}&0.4804(0.0059)&0.7659(0.0178)& 0.3663(0.0040)&0.4530(0.0132)&0.3053(0.0096) \\
\hline
{RMKM}&0.4514(0.0409)&0.7491(0.0277)&0.3236(0.0376)&0.4220(0.0197)&0.2865(0.0429) \\
\hline
{MVSC}&0.3341(0.0102)&0.5387(0.0271)&0.2427(0.0130)&0.1938(0.0185)&0.1242(0.0140) \\
\hline
MNMF&0.4414(0.0303)&0.7587(0.0330)&0.3115(0.0262)&0.4111(0.0175)& 0.3456(0.0576)\\
\hline
AMGL&0.6422(0.0139)&0.6638(0.0125)&0.6219(0.0164)&0.5711(0.0149)&0.4295(0.0208)\\					
\hline
{MVEC}&0.5862(0.0244)&0.4523(0.0114)&\textbf{0.8363(0.0654)}&\textbf{0.5924(0.0285)}&0.4318(0.0409)\\
\hline
{mPAC}&\textbf{0.6763}&0.6306& \textbf{0.7292}&\textbf{0.5741}&\textbf{0.4963}\\
\hline
{PFSC}& \textbf{0.7627(0.0117)}&\textbf {0.8687(0.0656)}&0.6836(0.0280)& 0.4388(0.0119)&\textbf{0.5832(0.0035)}\\

\hline
\end{tabular}}
\end{center}
\end{table}

\begin{table}[t]
\begin{center}
\renewcommand{\arraystretch}{1.1}
\caption{Clustering performance on Caltech20 data. }
\label{Caltech20} \scalebox{0.75}{
\begin{tabular}{|c|c|c|c|c|c|c}
\hline
{Method} & {F-Score}& {Precision} & {Recall}&{NMI}&{Adj-RI}\\
\hline

{KM}&0.3697(0.0071)&0.6235(0.0212)&0.2583(0.0095)&0.5578(0.0133)&0.2850(0.0063)\\

\hline
{Co-train}&0.3750(0.0287)&0.6375(0.0253)&0.2749(0.0238)&0.4895(0.0117)&0.3085(0.0281)\\

\hline
{Co-reg}&0.3719(0.0087)&0.6245(0.0137)&0.2882(0.0070)&0.5615(0.0042)&0.2751(0.0084)\\

\hline
{MKKM}&0.3583(0.0114)&\textbf{0.6724(0.0158)}&0.2865(0.0092)&0.5680(0.0142)&0.3039(0.0110) \\

\hline
{RMKM}&0.3955(0.0113)&0.6307(0.0144)&0.2712(0.0096)&\textbf{0.5899(0.0092)}&0.2952(0.0112)\\

\hline
MVSC& \textbf{0.5417(0.0239)}&0.4100(0.0245)& \textbf{0.7994(0.0110)}&0.4875(0.0113)&0.3800(0.0246)\\

\hline		
	MNMF&0.3643(0.0157)& \textbf{0.6509(0.0119)}&0.2530(0.0136)&0.5367(0.0132)&0.3128(0.0042)\\
	
	\hline
	AMGL&0.4017(0.0248)&0.3503(0.0479)&0.4827(0.0450)&0.5656(0.0387)&0.2618(0.0453)\\	

\hline
{MVEC}& 0.5229(0.0377)&0.4366(0.0412)& 0.7187(0.0499)&0.5841(0.0114)& \textbf{0.4517(0.0416)}\\
\hline
{mPAC}& \textbf{0.5645}&0.4350& \textbf{0.8035}&\textbf{0.5986}&\textbf{0.5083}\\
\hline
{PFSC}& 0.5163(0.0396)& 0.4337(0.0492)& 0.6432(0.0402)& 0.5790(0.0178)&0.4437(0.0415)\\

\hline
\end{tabular}}
\end{center}
\end{table}

Following \cite{cai2013multi}, we pre-process the data based on normalization, i.e., all values range from -1 to 1. For a fair comparison, we adopt five popular evaluation metrics as shown in \cite{kumar2011co}: F-Score, Precision, Recall, normalized mutual information (NMI) and Adjusted Rand Index (Adj-RI). For all of them, a larger value indicates a better clustering performance. For all methods, we either adopt the default parameter values given by the respective authors or tune them to achieve the best performance. Each method is repeated 10 times and the mean and standard deviation (std) values are reported.

\subsection{Experimental Result}
Tables \ref{Caltech7}-\ref{Handwritten} show the clustering results on the three benchmark data sets. For each measure, the best two methods are highlighted in boldface. In most cases, our proposed method achieves the best clustering performance in comparison with other state-of-the-art multi-view clustering methods. In particular,  
\begin{enumerate}
\item Our proposed PFSC often performs much better than current MVSC algorithm. Recall that both MVSC and PFSC learn multiple graphs. MVSC forces each graph to generate the same cluster, which is too restrictive considering the heterogeneous nature of views. PFSC produces a partition for each graph and fuses them to reach a consensus which allows for greater flexibility to deal with heterogeneous data. 
\item With respect to AMGL, another representative spectral clustering based method, PFSC outperforms it considerably in most cases. Though it assigns a weight for each graph, a common clustering is still assumed for all graphs. Moreover, its weight is empirically determined, which might not be correct for complex data. These factors   make AMGL impossible to perform well in all cases.
\item Compared to the multi-view ensemble clustering method MVEC, our method PFSC performs better in most cases. As we mention previously, they are different in several key aspects that are responsible for our improvements.
\item In general, graph-based methods perform better than $k$-means and NMF approaches, which is due to the efficiency of graph representation for the data in a non-Euclidean space \cite{gao2019exploring}. Graphs also provide a convenient way to explore supplementary information from multiple views.
\item With respect to mPAC, our method gives competitive performance. In some cases, our performance is even better. 
\end{enumerate}


 \begin{figure*}[!htb]
\includegraphics[scale=0.25]{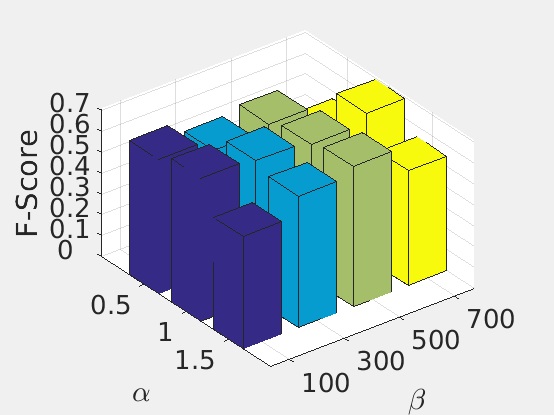}
\includegraphics[scale=0.25]{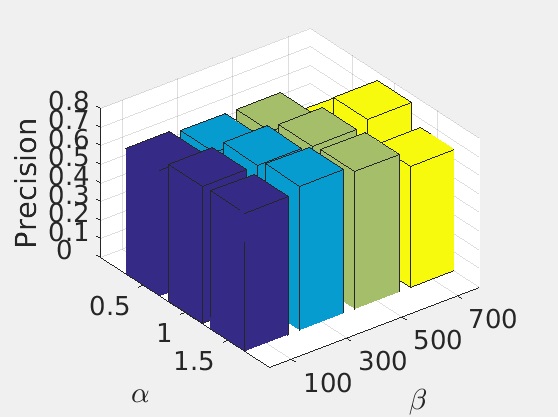}
\includegraphics[scale=0.25]{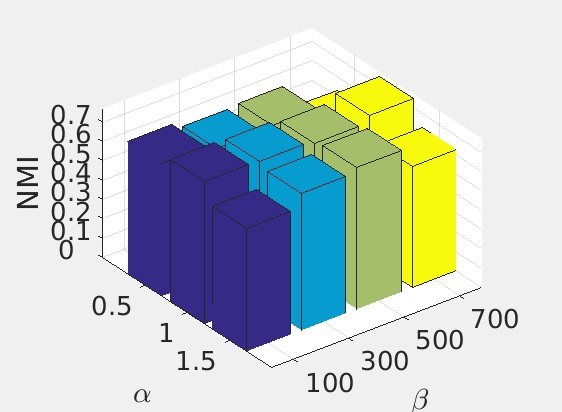}
\caption{Sensitivity analysis of parameters for our method over Caltech20 dataset evaluated with F-Score, Precision, and NMI.} \label{fig2}
\end{figure*}

To sum up, PFSC obtains competitive performance with respect to state-of-the-art techniques. This validates the effectiveness of partition fusion strategy. 
 \subsection{Parameter Sensitivity Analysis}
 There are two tuning parameters, $\alpha$ and $\beta$ in our model (\ref{final_method}). Taking Caltech20 and Caltech7 as examples, we empirically examine their influence on F-score, Precision, and NMI, as shown in Figures \ref{fig1} and  \ref{fig2}. It can be observed that the performance of PFSC is relatively stable for a wide range of parameters.

 \begin{table}[t]
\begin{center}
\renewcommand{\arraystretch}{1.1}
\caption{Clustering performance on Handwritten numerals data. }
\label{Handwritten} \scalebox{0.75}{
\begin{tabular}{|c|c|c|c|c|c|c}
\hline
{Method} & {F-Score}& {Precision} & {Recall}&{NMI}&{Adj-RI}\\
\hline

{KM}&0.6671(0.0105)&0.6550(0.0154)&0.6889(0.0180)&0.7183(0.0106)&0.6443(0.0122)\\

\hline
{Co-train}&0.6859(0.0172)&0.6634(0.0281)&0.7109(0.0252)&0.7222(0.0149)&0.6498(0.0227) \\

\hline
{Co-reg}&0.6840(0.0269)&0.6360(0.0336)&0.6413(0.0198)&0.7583(0.0197)&0.6266(0.0314) \\

\hline
{MKKM}&0.6756(0.0000)&0.6501(0.0000)&0.7050(0.0000)&0.7526(0.0000)&0.7009(0.0000) \\
\hline
{RMKM}&0.6542(0.0258)&0.6218(0.0350)&0.6915(0.0158)&0.7431(0.0209)&0.6013(0.0300) \\

\hline
{MVSC}&0.6753(0.0335)&0.6193(0.0537)&\textbf{0.7537(0.0215)}&0.7566(0.0186)&0.6079(0.0419) \\

\hline
MNMF&  0.7068(0.0272)& 0.6957(0.0294)&0.7183(0.0250)&0.7431(0.0227)&0.6407(0.0056)\\

\hline
AMGL&\textbf{0.7404(0.1070)}&0.6650(0.1372)&\textbf{0.8457(0.0560)}&\textbf{0.8392(0.0543)}&\textbf{0.7066(0.1235)}\\

\hline
{MVEC}&0.7196(0.0313)&\textbf{0.8082(0.0157)}&0.6501(0.0437)&\textbf{0.8166(0.0142)}&0.6847(0.0361)\\
\hline
{mPAC}& \textbf{0.7473}&0.7348&0.7200&0.7370&\textbf{0.7069}\\
\hline
{PFSC}& 0.7263(0.0249)&\textbf {0.7549(0.0176)}&  0.7001(0.0326)&0.7666(0.0148)& 0.6948(0.0281)\\

\hline
\end{tabular}}
\end{center}
\end{table}
%

\section{Experiments on Noisy Data}
\label{noise}

 \begin{figure}[!htb]
 \centering
\includegraphics[width=.45\textwidth]{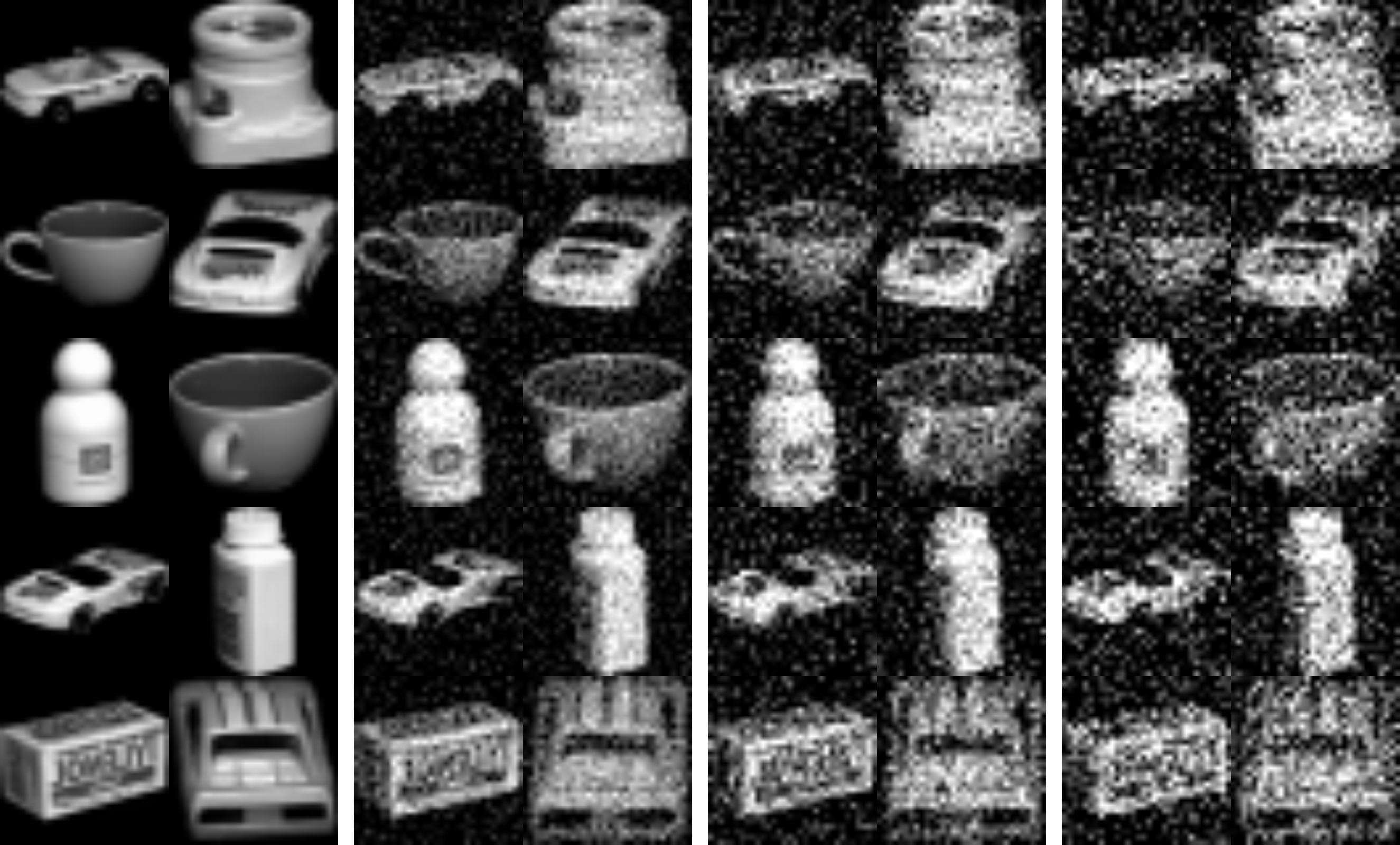}
\caption{The first column is some COIL20 images. They are corrupted by Gaussian noise with different variance: 0.01, 0.03, 0.05 (from 2nd to 4th column).} \label{gauss}
\end{figure}
 \begin{figure}[!htb]
 \centering
\includegraphics[width=.45\textwidth]{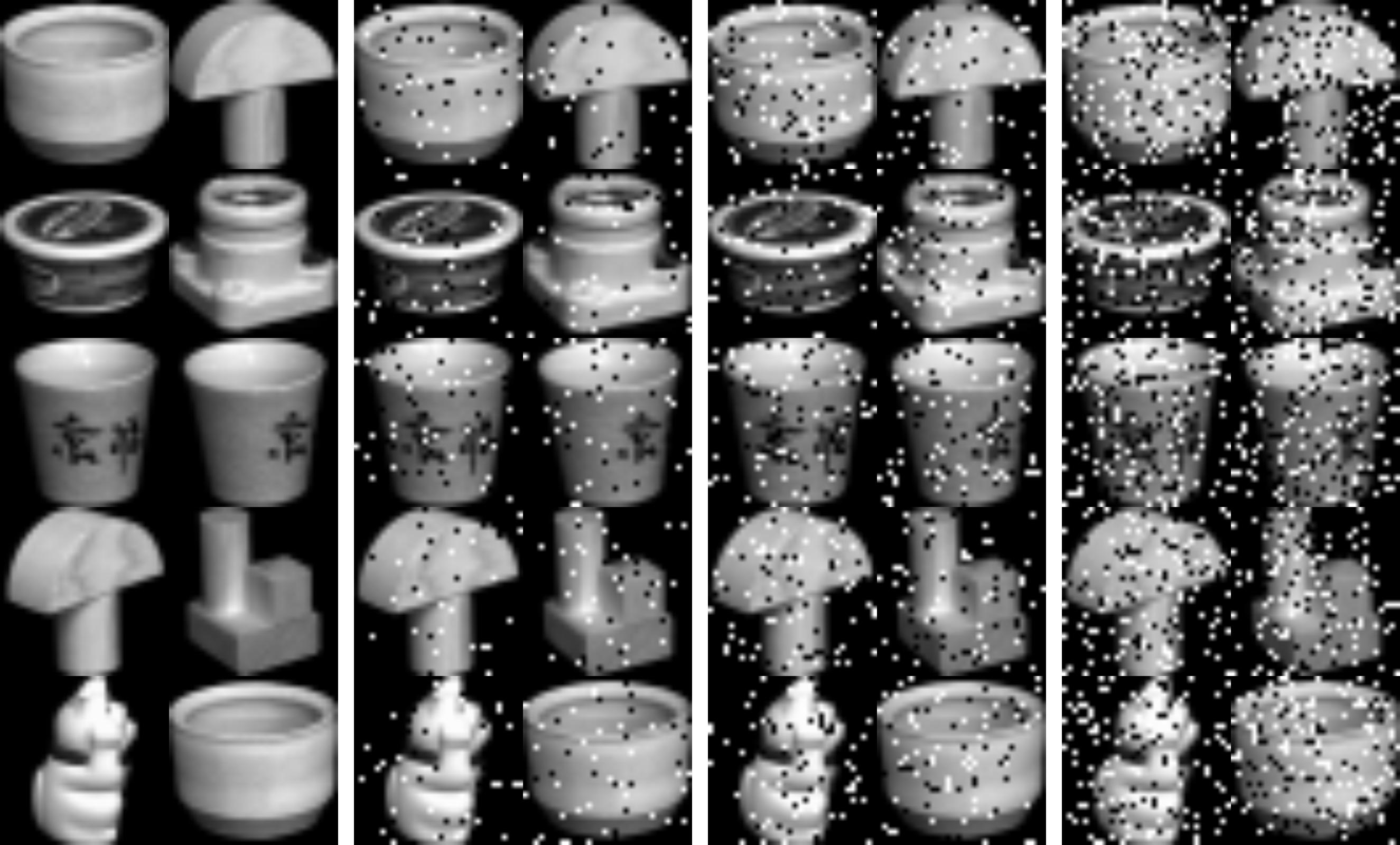}
\caption{The first column is some COIL20 images. They are corrupted by Salt \& Pepper noise with different density: 0.05, 0.1, 0.2 (from 2nd to 4th column).} \label{salt}
\end{figure}
In practice, images are usually liable to suffer from noises. To examine the robustness of our model, we contaminate clean images by adding two commonly seen noises: Gaussian noise and Salt \& Pepper noise. We first construct a single-view data set by randomly sampling 24 images for each class from COIL20 \cite{nene1996columbia} which is a data set of 20 toy images. This results in a data set with size $1024\times 480$. Then, we add three levels of noise to obtain three noise data. Based on them, we form a three-view data set. In specific, we corrupt the images with Gaussian noise with zero mean and variance 0.01, 0.03, and 0.05; Salt \& Pepper noise with density 0.05, 0.1, and 0.2. Some sample images are displayed in Figures \ref{gauss} and \ref{salt}. We can observe that those images are heavily corrupted.

Here, we run $k$-means on all partitions from our model (\ref{final_method}), i.e., $F_1$, $F_2$, $F_3$, $Y$, and report their results as PFSC-$F_1$, PFSC-$F_2$, PFSC-$F_3$, PFSC-$Y$, respectively. We compare PFSC with two most relevant gragh-based multi-view clustering method: MVSC and AMGL. In addition, we choose them since they produce very competitive performance in Tables \ref{Caltech20}-\ref{Handwritten}. In fact, MVSC is a robust learning algorithm, in which an error variable $E^v$ is purposely introduced to characterize noise for each view, and then $\ell_1$-norm regularization is imposed to $E^v$ to enforce the sparsity for the outlying entries. 
\begin{table}[t]
\begin{center}
\renewcommand{\arraystretch}{1.8}
\caption{Clustering performance on noisy data. }
\label{noise} \scalebox{0.60}{
\begin{tabular}{|c|c|c|c|c|c|c|}
\hline
{Noise}&{Method} & {F-Score}& {Precision} & {Recall}&{NMI}&{Adj-RI}\\
\hline

\multirow{6}*{Gaussian}&\multirow{1}*{MVSC}&0.3467(0.0321)&0.5194(0.0195)&0.2294(0.0347)&0.5615(0.0229)&0.2677(0.0373)\\

\cline{2-7}
&\multirow{1}*{AMGL}&0.4240(0.0734)&0.3076(0.0696)&\textbf{0.7036(0.0340)}&\textbf{0.7754(0.0343)}&0.3822(0.0820)\\

\cline{2-7}
&\multirow{1}*{PFSC-$F_1$}&0.5156(0.0528)&0.6439(0.0366)&0.4469(0.0629)&0.7506(0.0239)&0.4908(0.0576)\\

\cline{2-7}
&\multirow{1}*{PFSC-$F_2$}&0.5200(0.0558)&0.6595(0.0364)&0.4312(0.0632)&0.7525(0.0279)&0.4902(0.0605)\\

\cline{2-7}

&\multirow{1}*{PFSC-$F_3$}&0.5086(0.0267)&\textbf{0.6613(0.0232)}&0.4143(0.0338)&0.7483(0.0163)&0.4777(0.0293)\\

\cline{2-7}
&\multirow{1}*{PFSC-Y}&\textbf{0.5226(0.0395)}&0.6257(0.0216)&0.4508(0.0529)&0.7590(0.0200)&\textbf{0.4943(0.0431)}\\

\hline


\multirow{6}*{Salt \& Pepper}&\multirow{1}*{MVSC}&0.3510(0.0363)&0.5034(0.0240)&0.2710(0.0397)&0.5709(0.028)&0.3076(0.0408)\\

\cline{2-7}
&\multirow{1}*{AMGL}&0.3954(0.0464)&0.2800(0.0451)&\textbf{0.6836(0.0189)}&0.7539(0.0175)&0.3509(0.0520)\\	

\cline{2-7}
&\multirow{1}*{PFSC-$F_1$}&0.5065(0.0459)&0.6042(0.0336)&0.4290(0.0566)&0.7493(0.0230)&0.4765(0.0502)\\

\cline{2-7}
&\multirow{1}*{PFSC-$F_2$}&0.5218(0.0583)&0.6105(0.0489)&0.4575(0.0651)&0.7516(0.0291)&0.4939(0.0629)\\

\cline{2-7}
&\multirow{1}*{PFSC-$F_3$}&0.5016(0.0398)&0.6118(0.0393)&0.4284(0.0539)&0.7443(0.0212)&0.4717(0.0436)\\

\cline{2-7}
&\multirow{1}*{PFSC-Y}&\textbf{0.5325(0.0345)}&\textbf{0.6190(0.0222)}&0.4683(0.0429)&\textbf{0.7593(0.0175)}&\textbf{0.5055(0.0373)}\\	

\hline

\end{tabular}}
\end{center}
\end{table}
 \begin{figure*}[!htbp]
\subfloat[The 1st View]{\includegraphics[width=.48\textwidth]{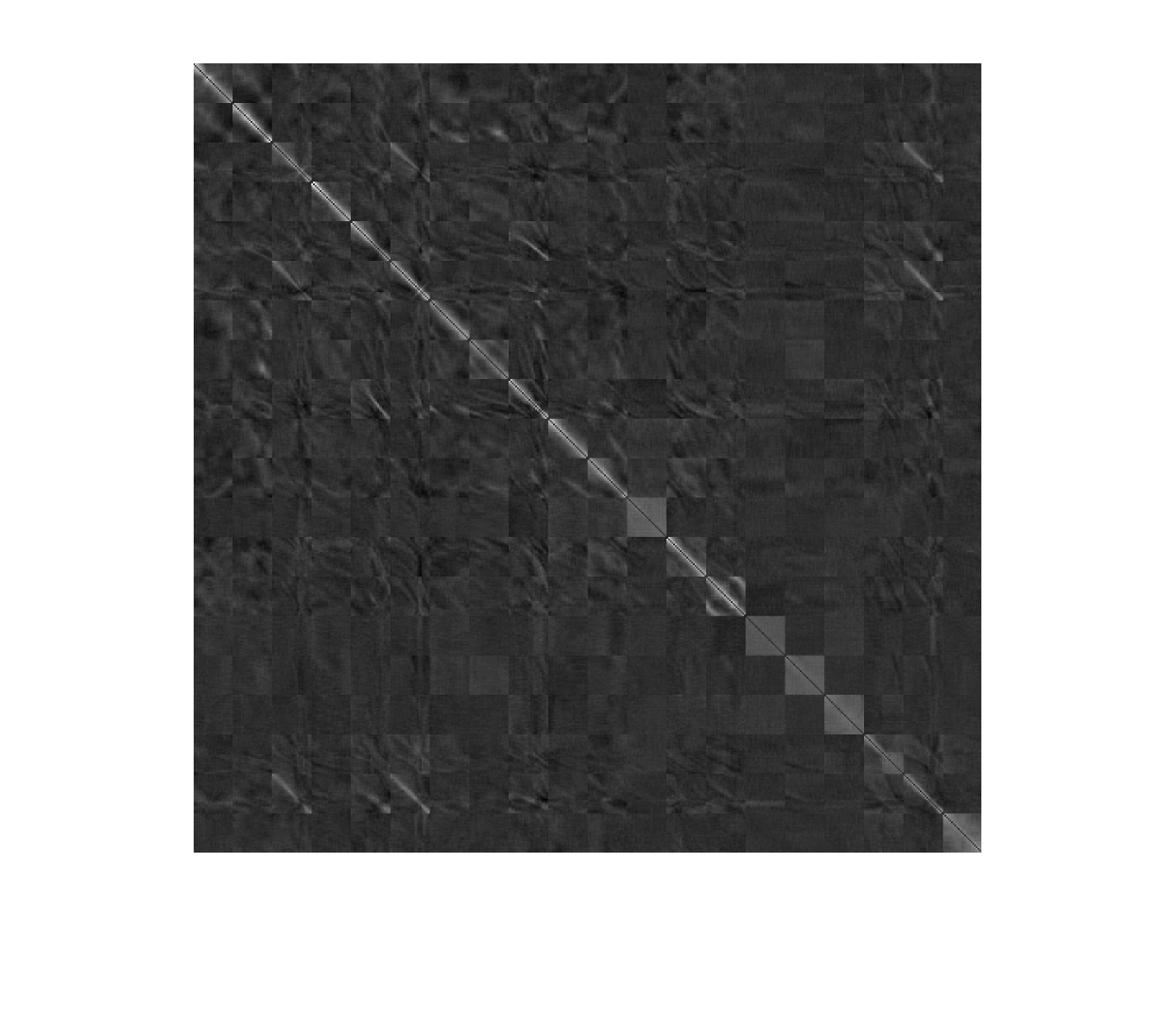}}
\subfloat[The 2nd View]{\includegraphics[width=.48\textwidth]{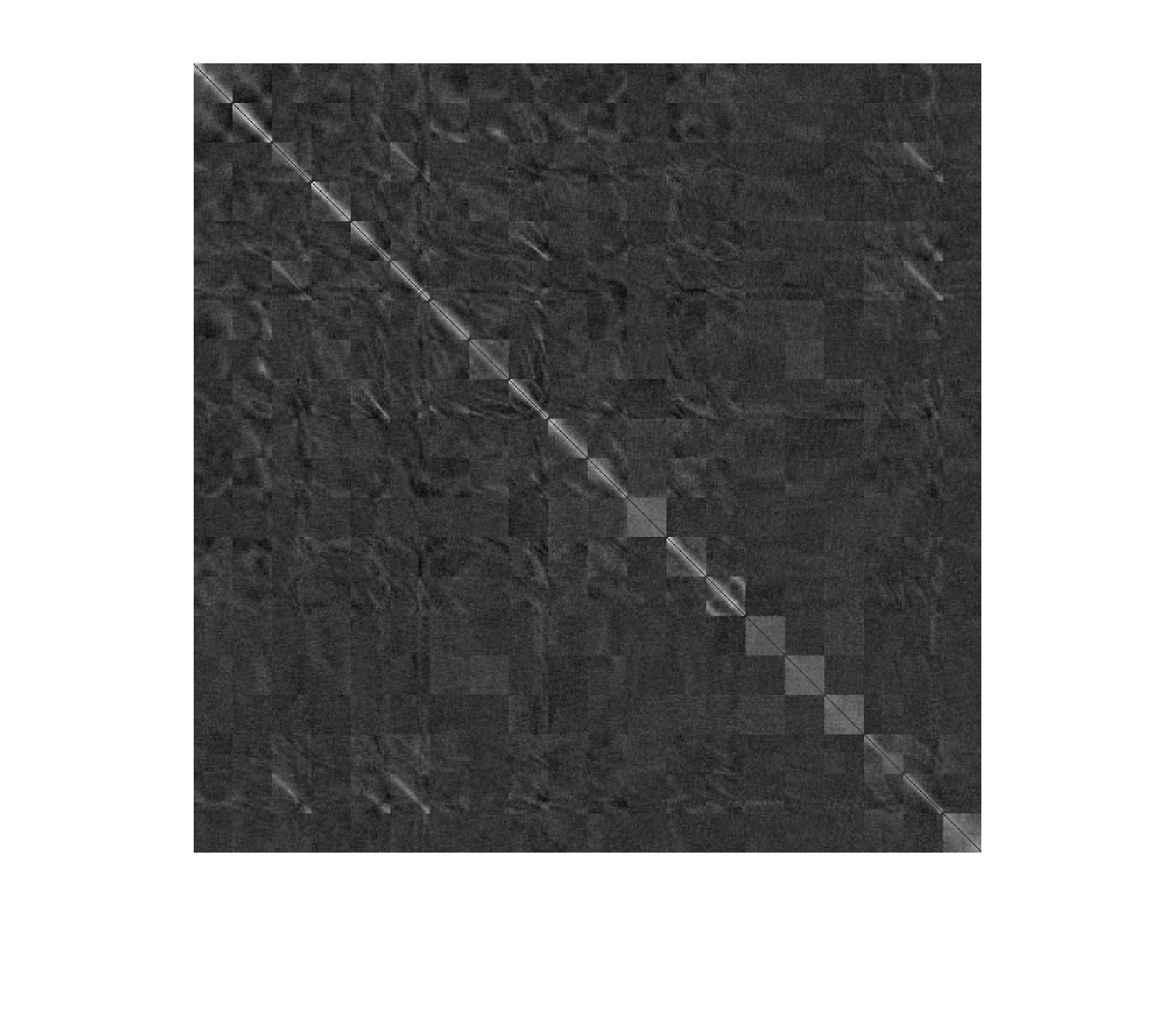}}\\
\subfloat[The 3rd View]{\includegraphics[width=.48\textwidth]{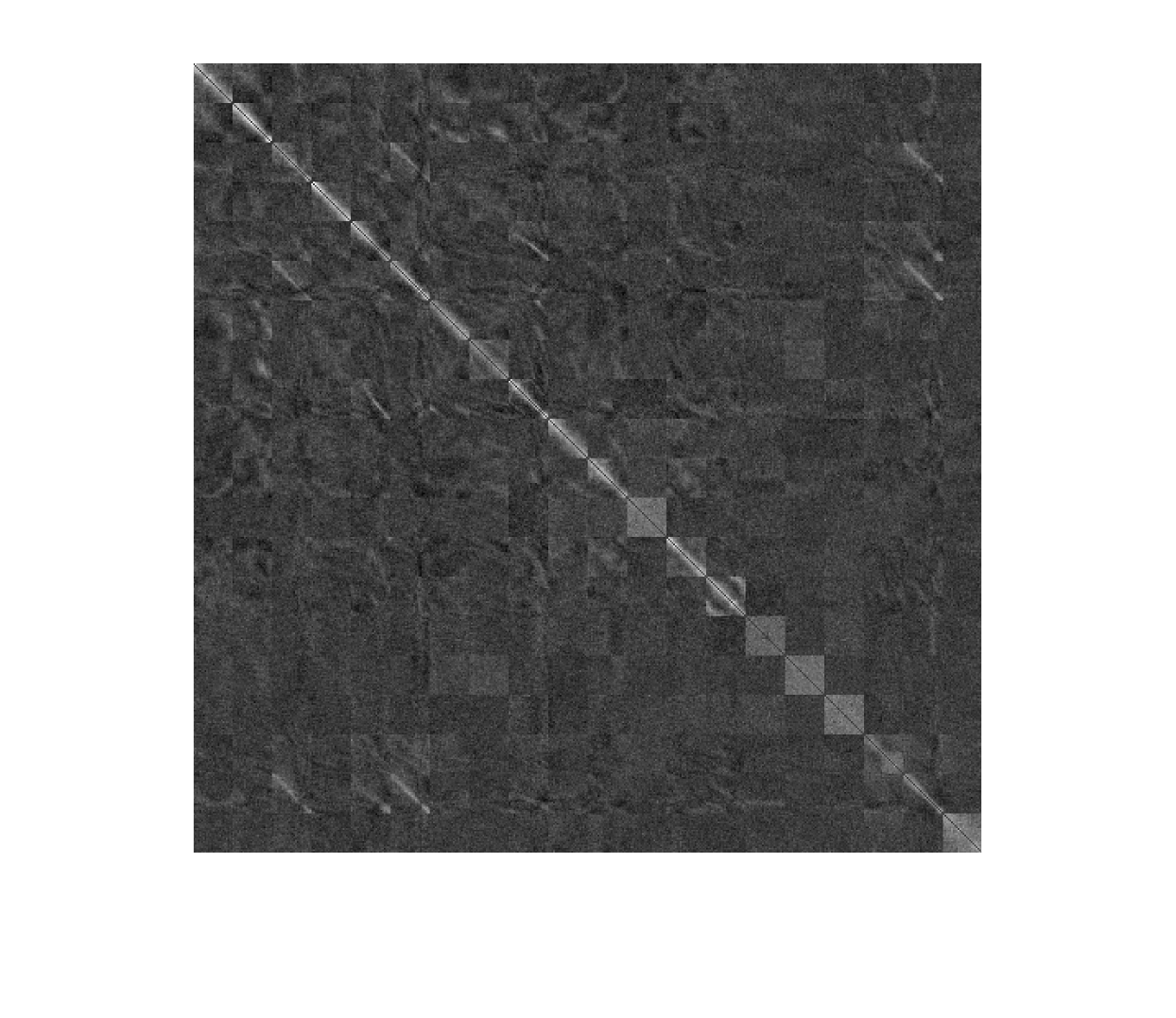}}

\caption{The learned graphs $S$ corrupted by Gaussian noise.} \label{sg}
\end{figure*}

 \begin{figure*}[!htbp]
\subfloat[Partition $F_1$]{\includegraphics[width=.49\textwidth]{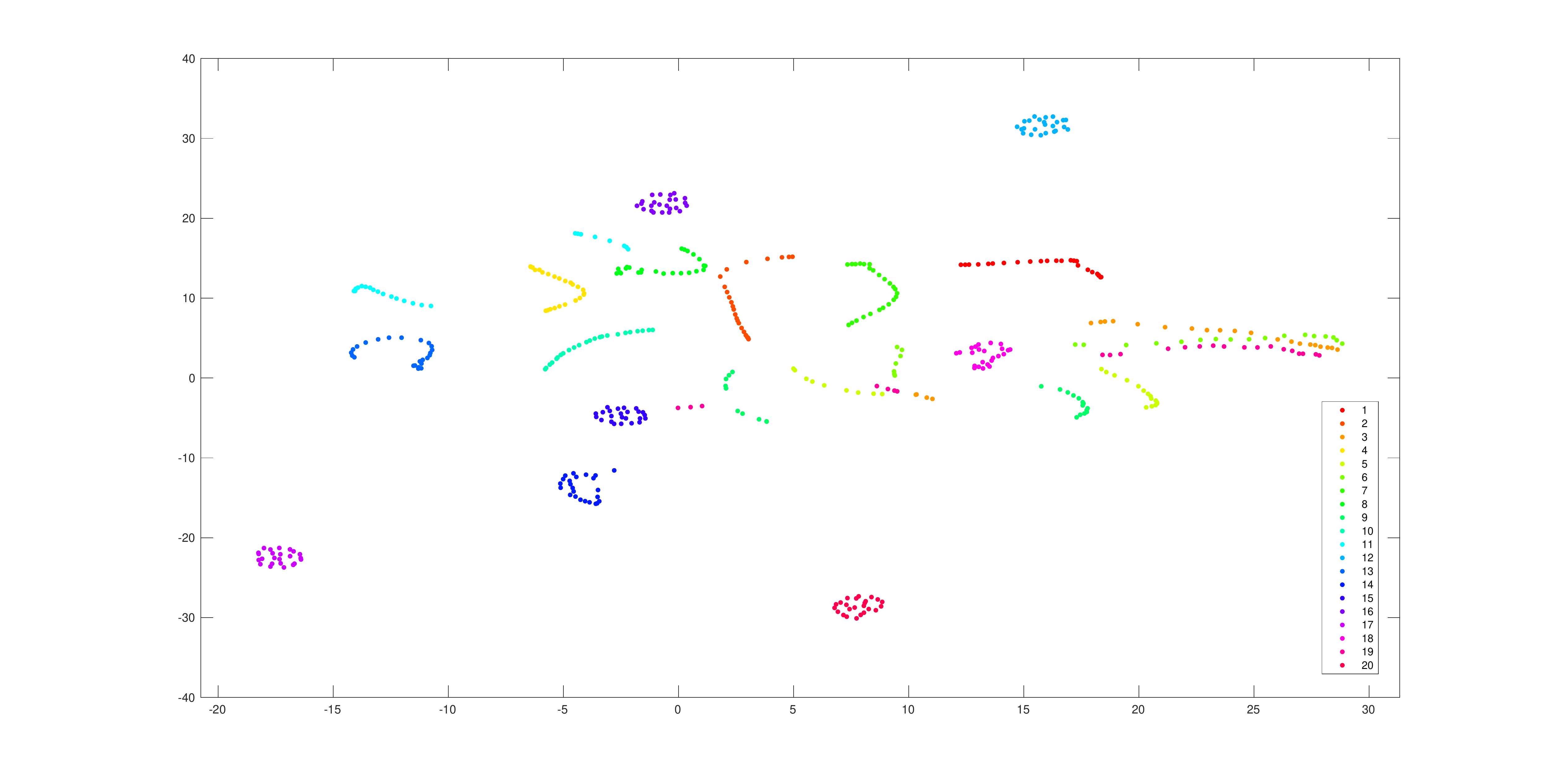}}
\subfloat[Partition $F_2$]{\includegraphics[width=.49\textwidth]{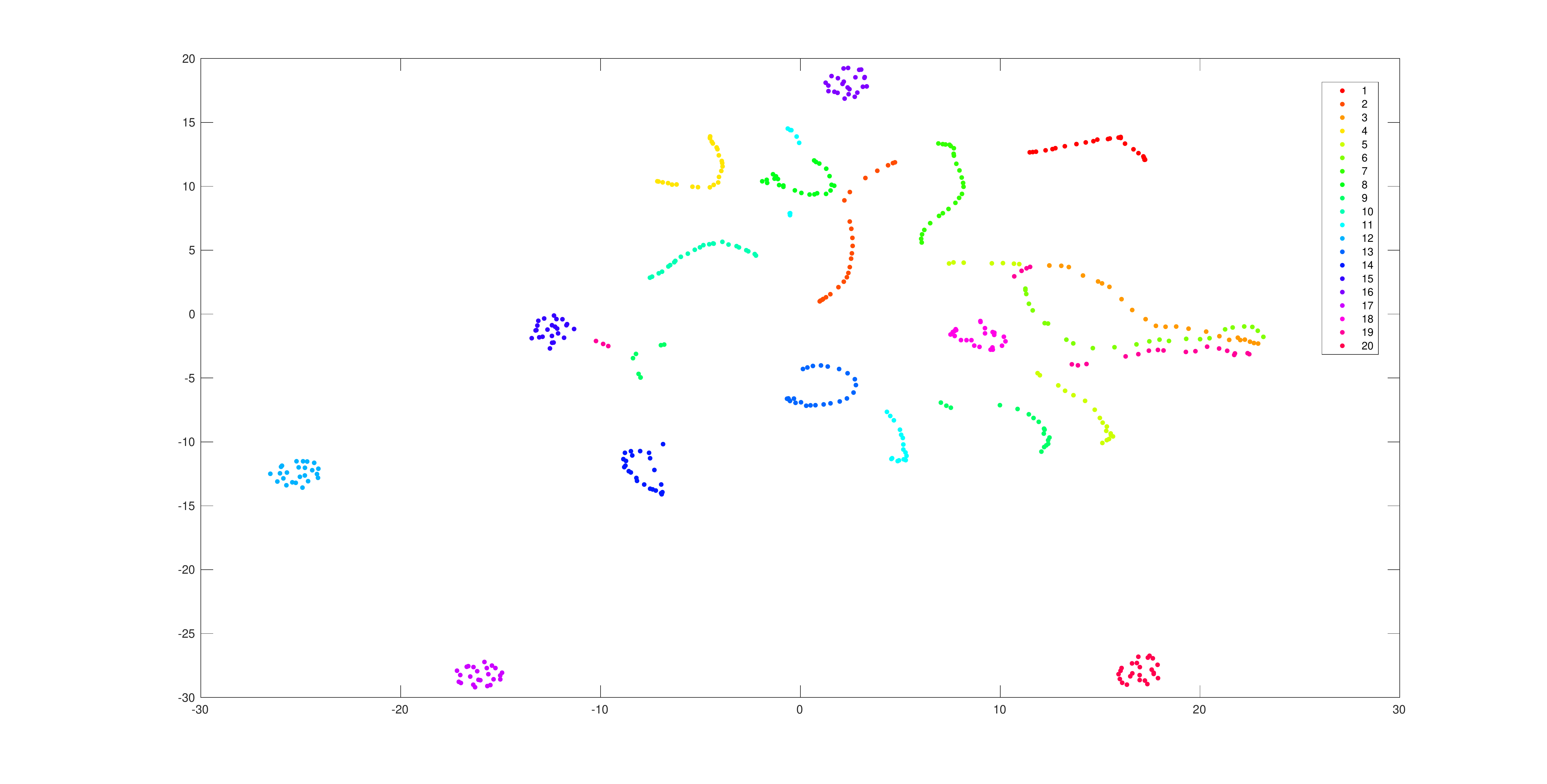}}\\
\subfloat[Partition $F_3$]{\includegraphics[width=.49\textwidth]{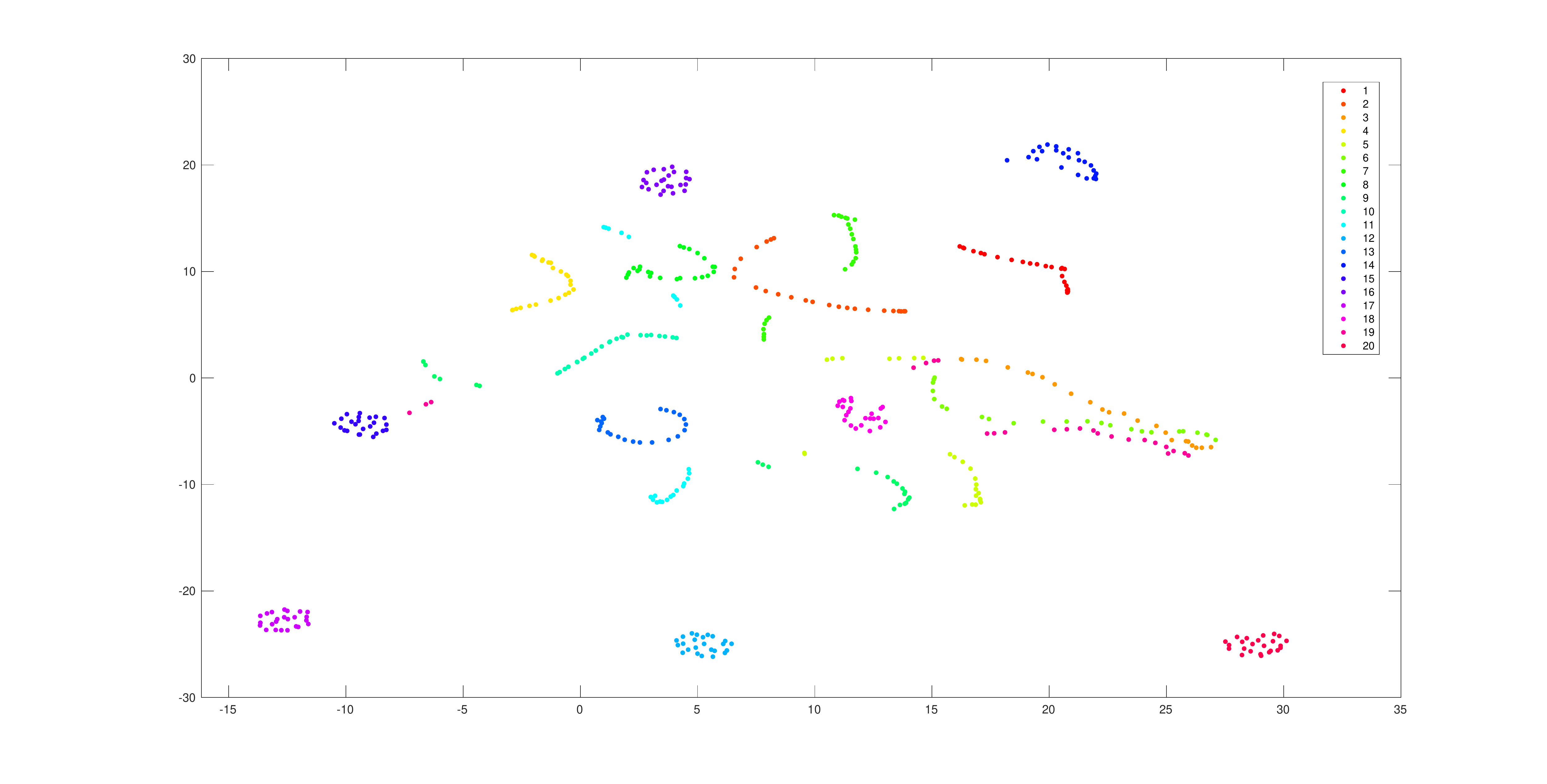}}
\subfloat[Final clustering Y]{\includegraphics[width=.49\textwidth]{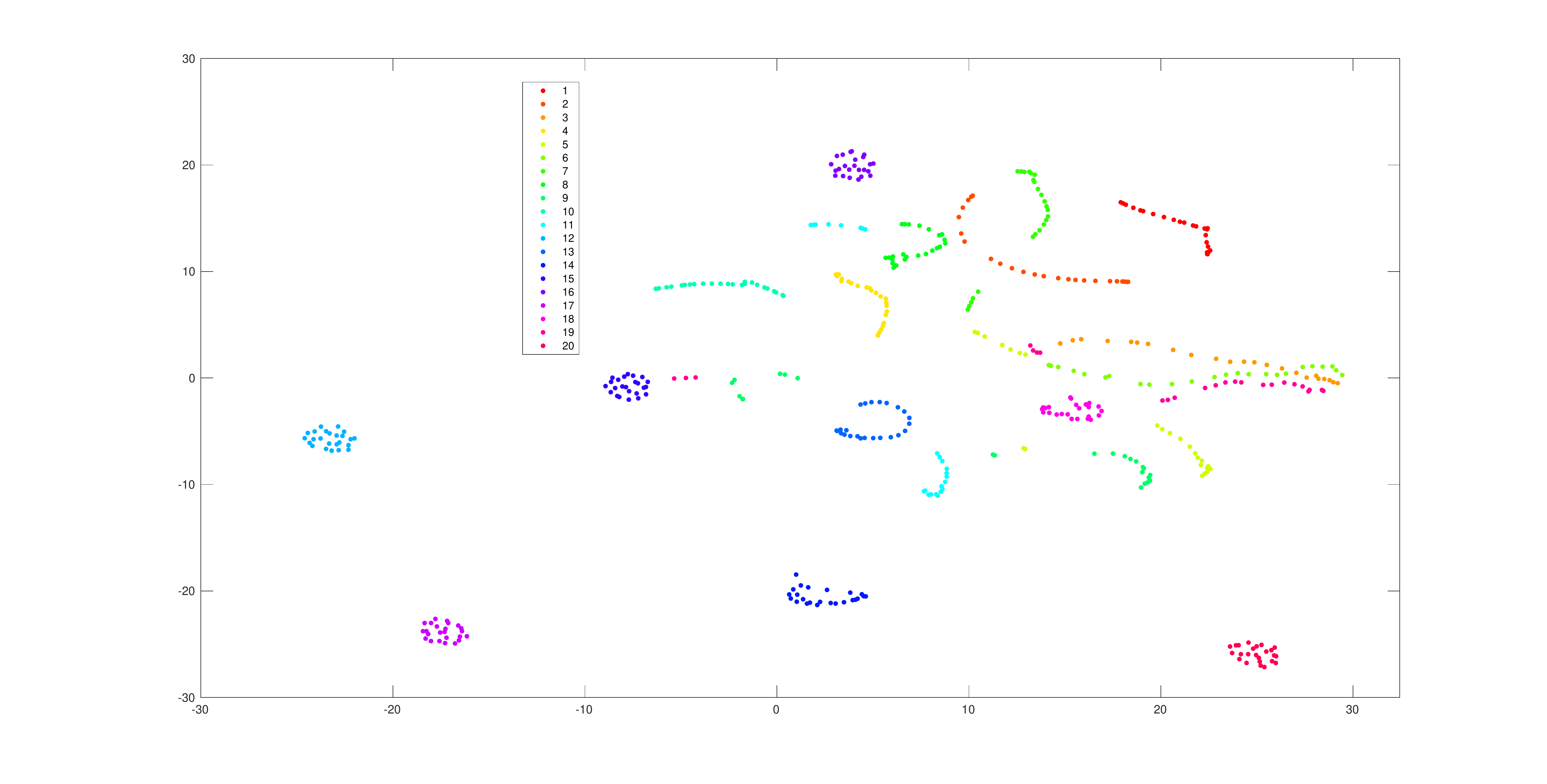}}
\caption{Visualization of partitions corrupted by Gaussian noise.} \label{pg}
\end{figure*}

 \begin{figure*}[!htbp]
\centering
\subfloat[The 1st View]{\includegraphics[width=.48\textwidth]{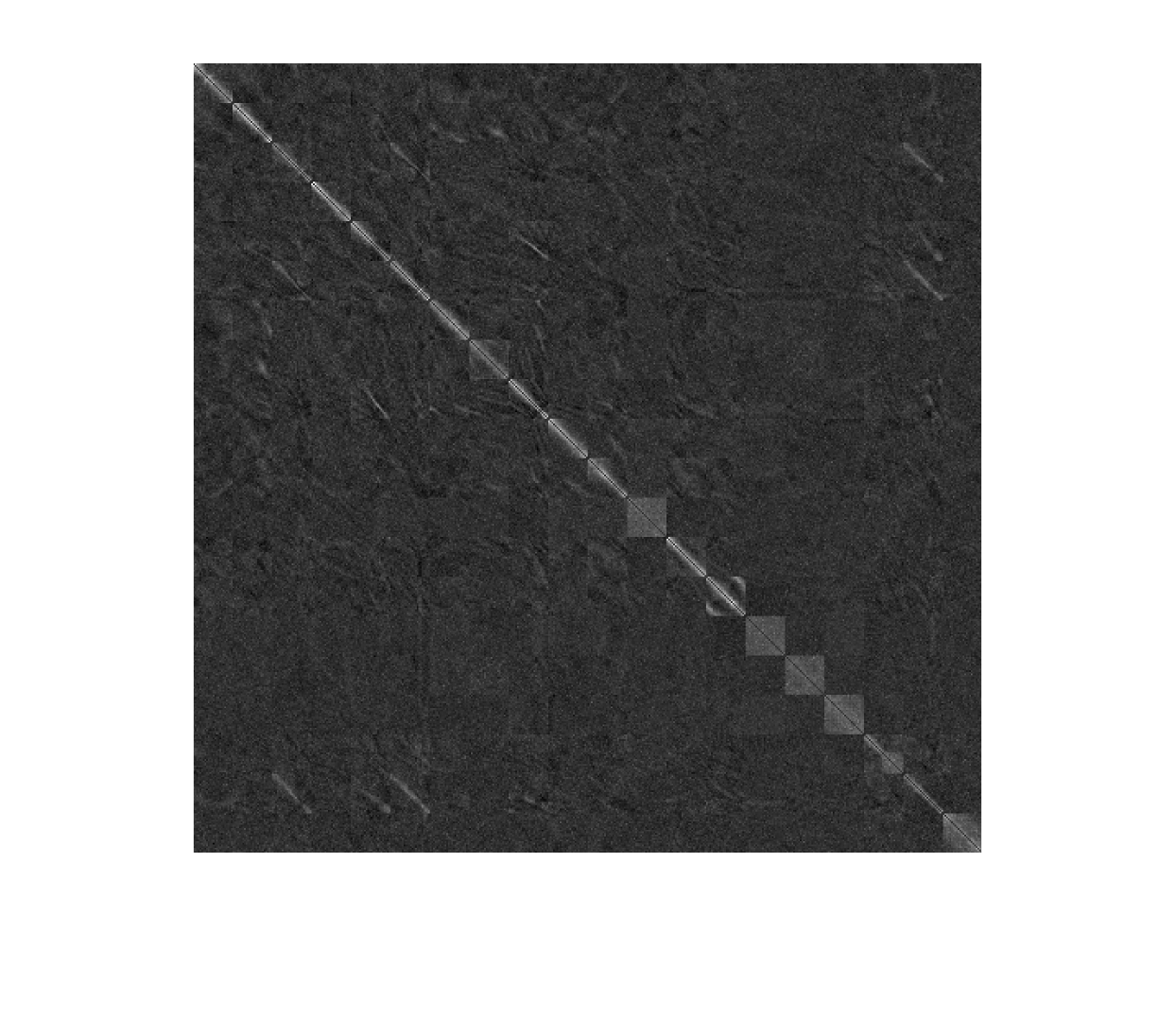}}
\subfloat[The 2nd View]{\includegraphics[width=.48\textwidth]{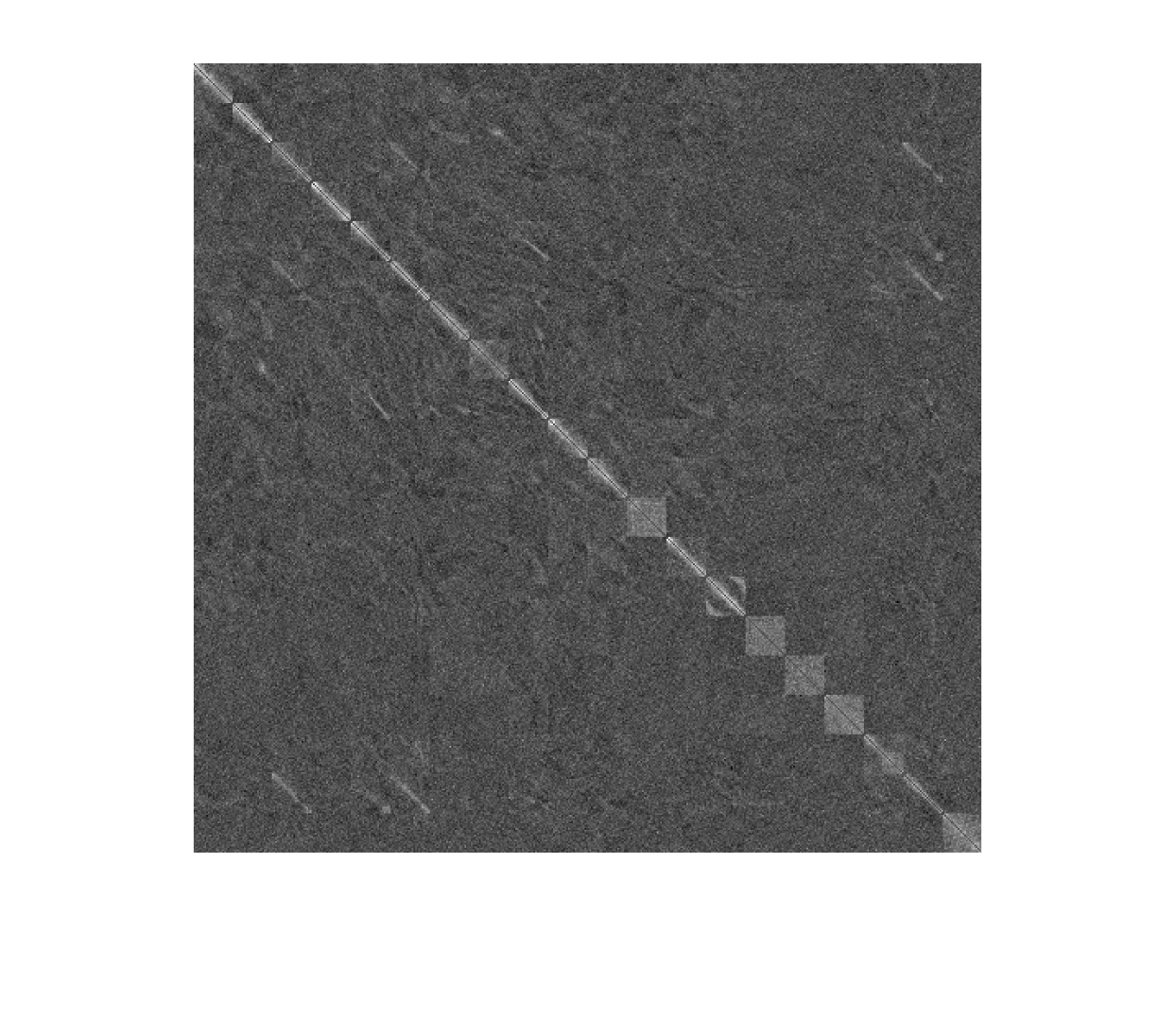}}\\
\subfloat[The 3rd View]{\includegraphics[width=.48\textwidth]{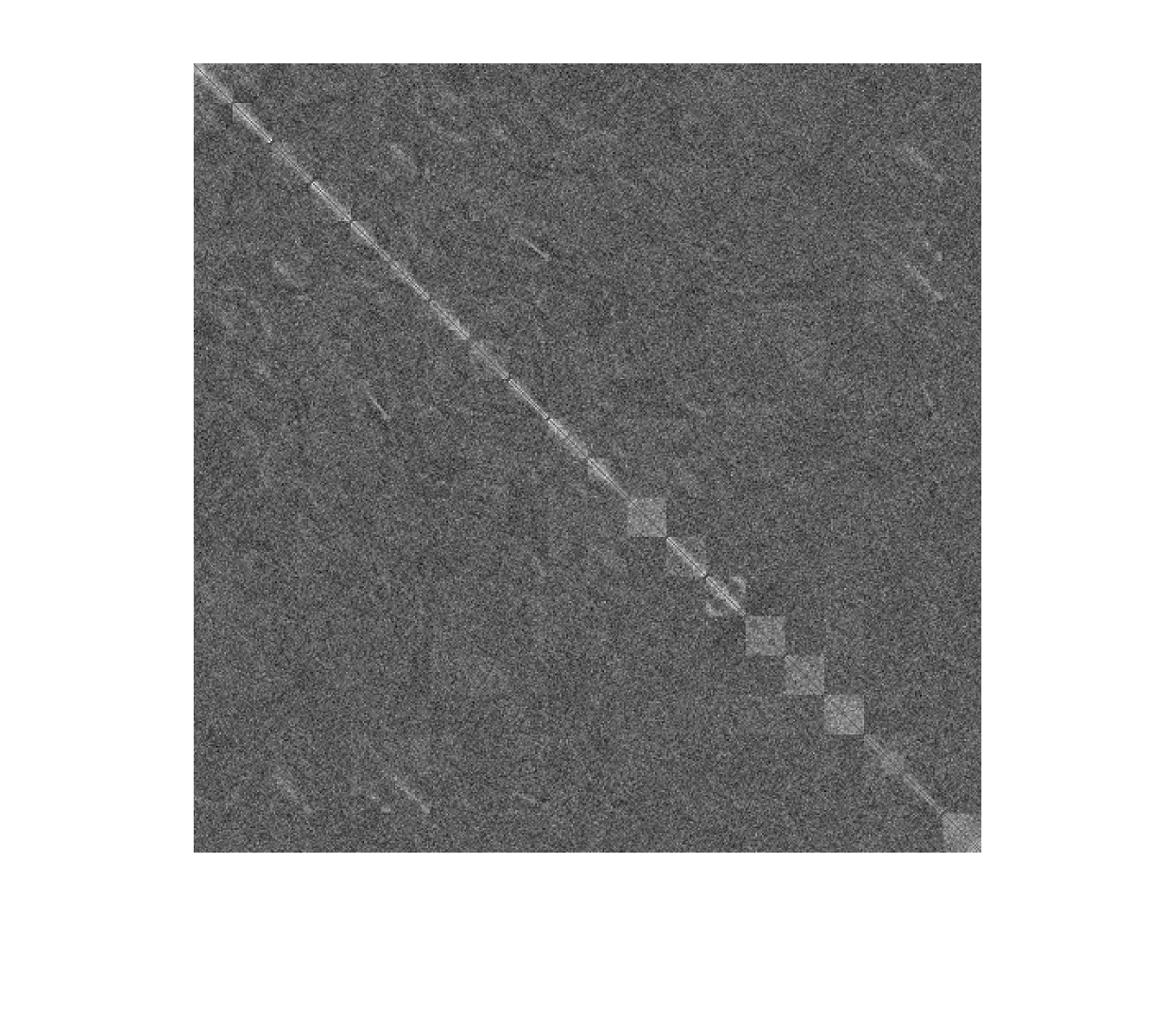}}

\caption{The learned graphs $S$ corrupted by Salt \& Pepper noise.} \label{ssa}
\end{figure*}
 \begin{figure*}[!htbp]
\centering
\subfloat[ Partition F1]{\includegraphics[width=.49\textwidth]{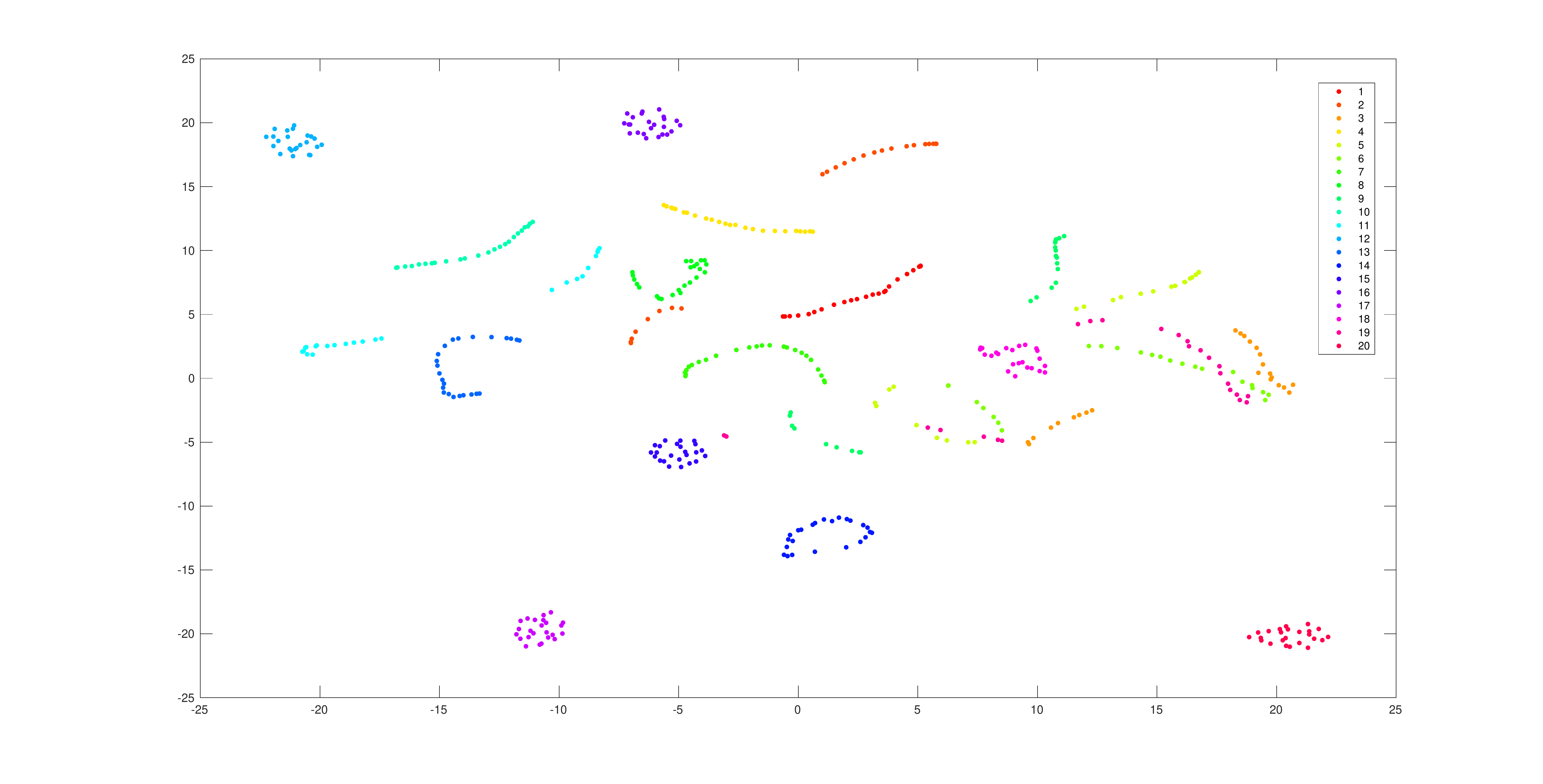}}
\subfloat[Partition F2]{\includegraphics[width=.49\textwidth]{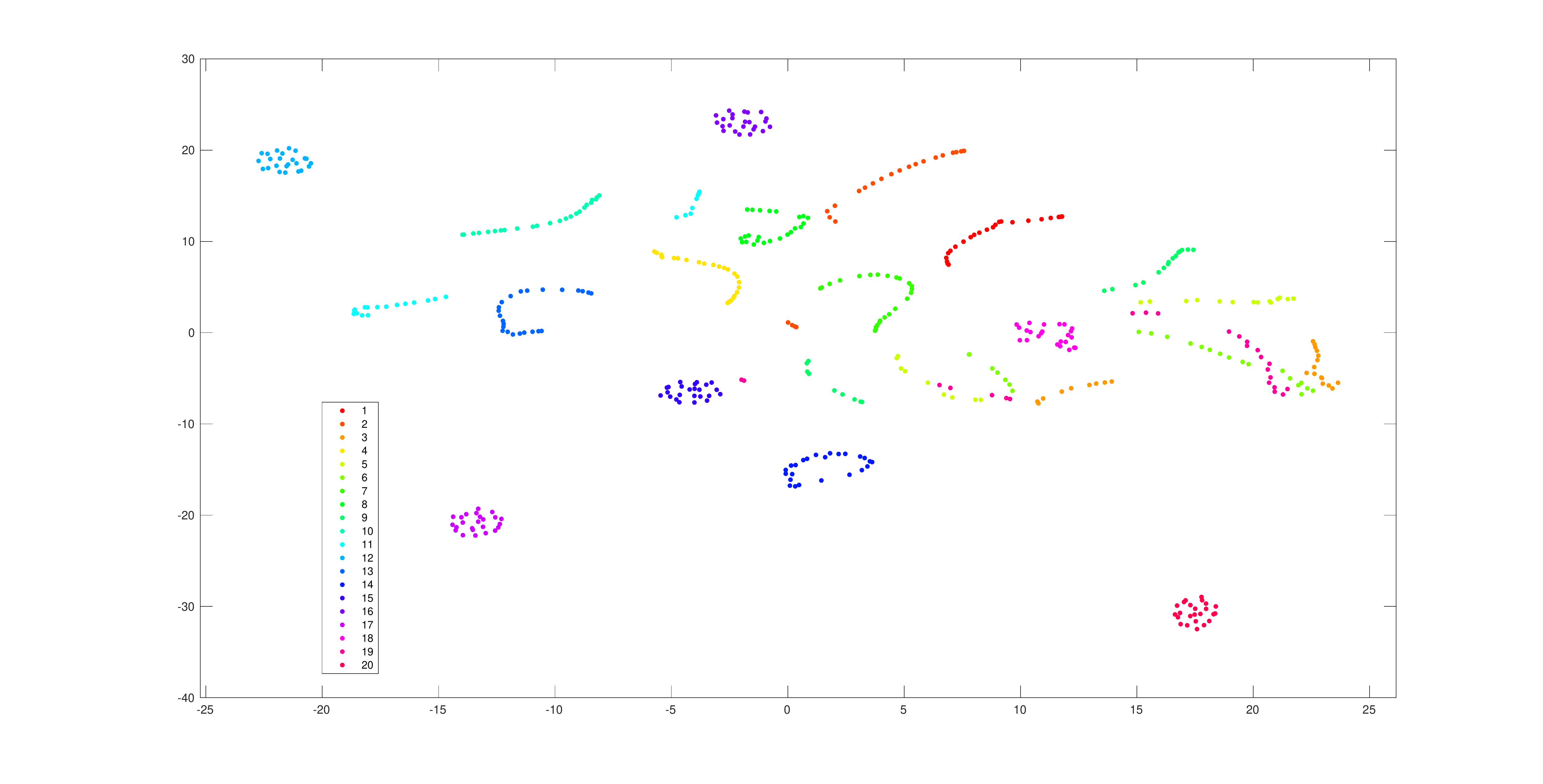}}\\
\subfloat[Partition F3]{\includegraphics[width=.49\textwidth]{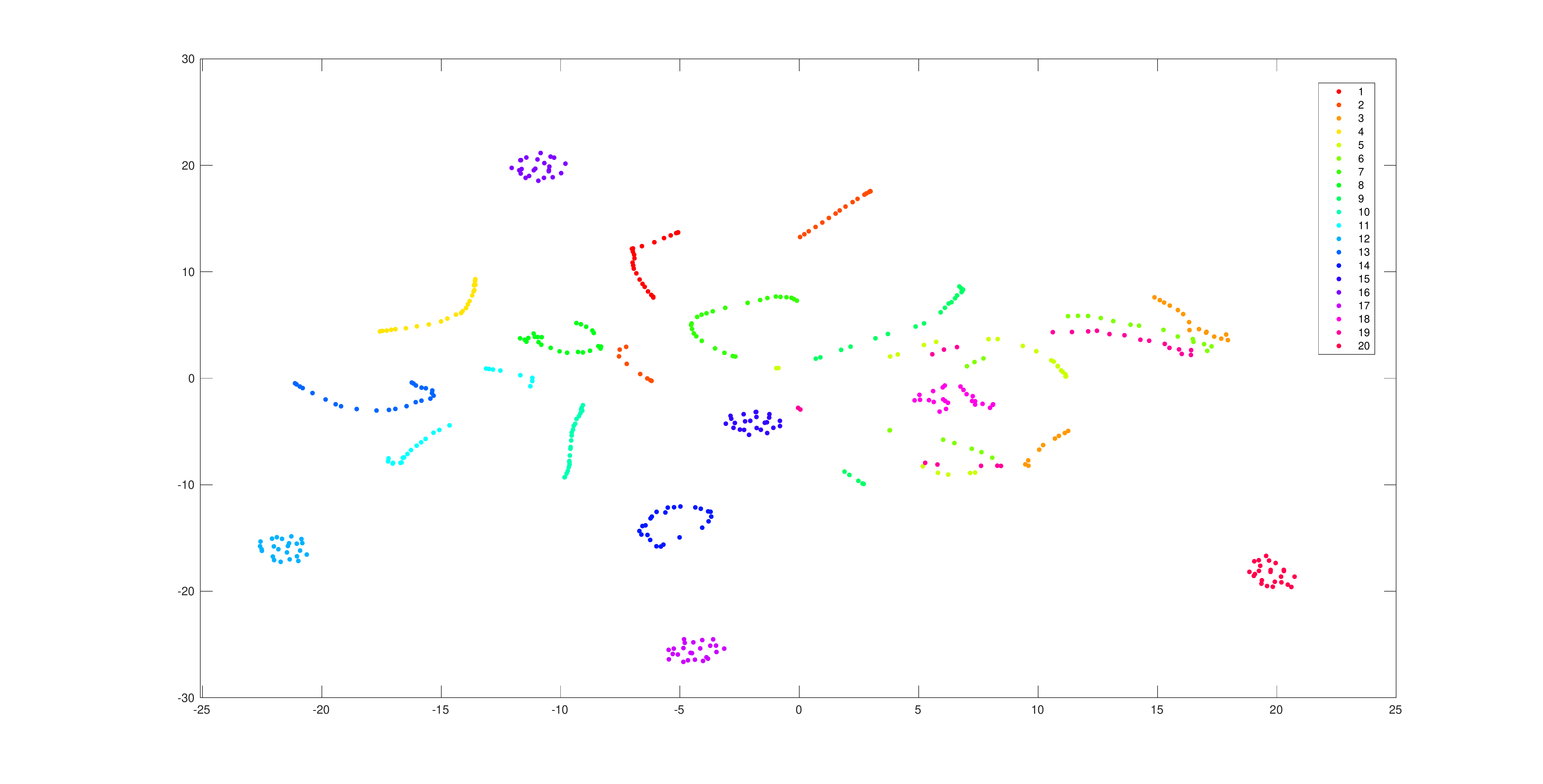}}
\subfloat[Final clustering Y]{\includegraphics[width=.49\textwidth]{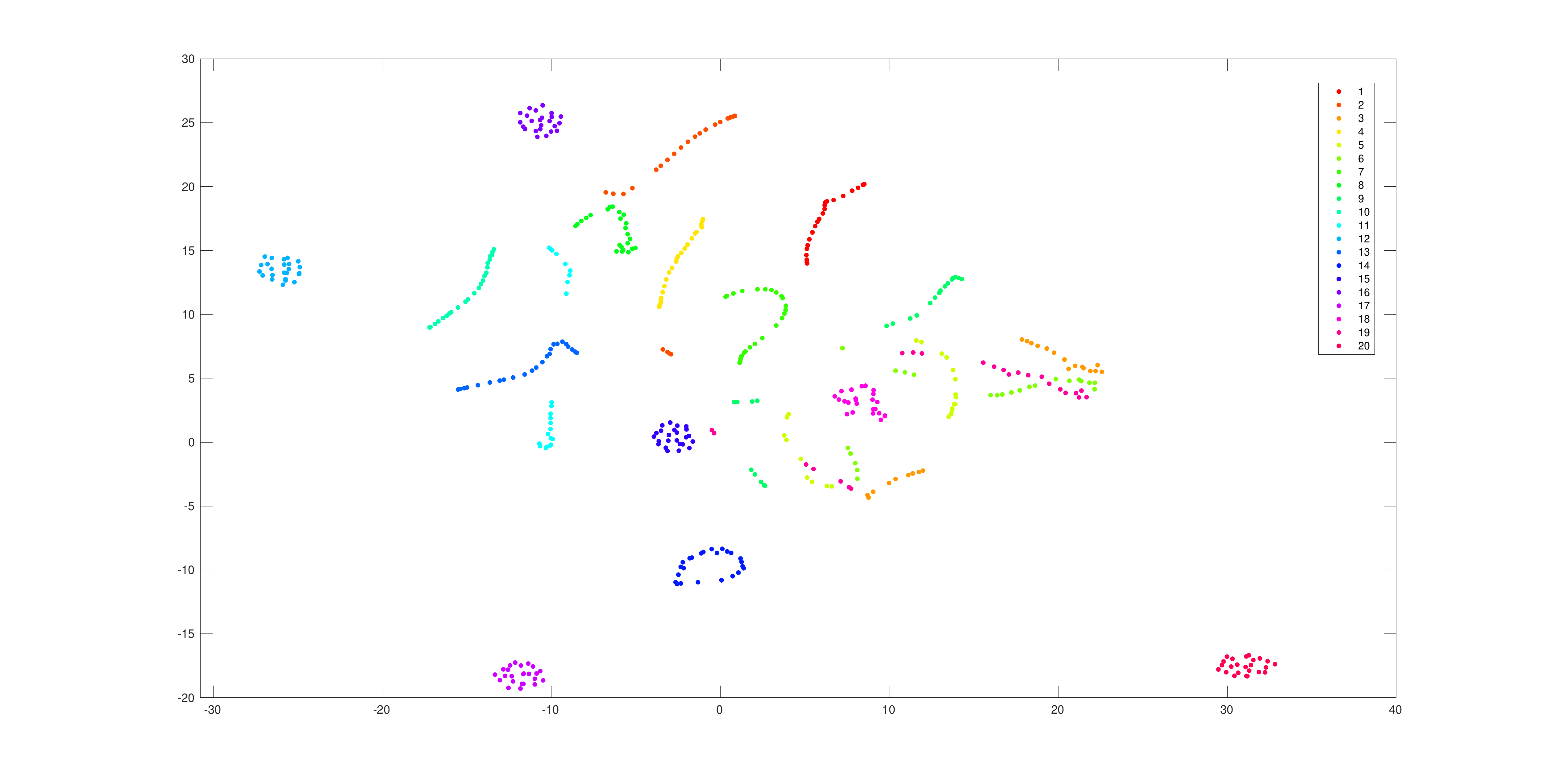}}
\caption{ Visualization of partitions corrupted by Salt \& Pepper noise.} \label{psa}
\end{figure*}
Table \ref{noise} summarizes the learning performances of the algorithms on noisy data, and it can be observed that PFSC-$Y$ outperforms MVSC and AMGL significantly by more than 10\% in terms of F-Score, Precision, and Adj-RI. Though MVSC is designed to resist noise, it is inferior to PFSC-$Y$ in all cases. This is due to the fact that MVSC treats each view equally and assumes that all graphs share the same partition matrix. To reach the overall minimum of objective function, the final cluster pattern could be deteriorated. On the other hand, both AMGL and PFSC-$Y$ adopt a weighting strategy, hence they can assign more weight to high quality view. Different from AMGL, which uses the same partition for all views, PFSC-$Y$ finds the final clustering through fusing multiple basic partitions. Since all views admit the same cluster pattern, it becomes easy for PFSC to search for the best one. 

Furthermore, we can observe that our final clustering PFSC-$Y$ often outperforms PFSC-$F_1$, PFSC-$F_2$, and PFSC-$F_3$. This validates the effectiveness of our fusion strategy. Based on those partitions from each single view, we can eventually achieve a better clustering. The learned graphs and partitions are also displayed in Figures \ref{sg}-\ref{psa}. To obtain Figures \ref{pg} and \ref{psa}, t-SNE is implemented \cite{maaten2008visualizing}. We can see that the learned graphs can not present the cluster structure. Ideally, they should be block-diagonal. On the other hand, the partitions can explicitly display the cluster structure. This demonstrates the robustness of our model.

\section{Conclusion}\label{conclude}
In this paper, we develop a novel multi-view subspace clustering method, aiming to exploring multi-view information in a partition space to enhance model robustness. The proposed model integrates graph learning, spectral clustering, weight learning, and partition fusion into a unified framework, in which each component is optimized. Experimental results on widely adopted benchmark data sets validate the effectiveness and robustness of the proposed method. 


%
%


\bibliographystyle{elsarticle-num}
\bibliography{ref}

\end{document}